\documentclass[conference,screen]{IEEEtran}

\usepackage{amsfonts}
\usepackage{mathtools} 
\usepackage[english]{babel}
\usepackage{amsmath,amssymb,amsthm}
\usepackage{tabularx}
\usepackage{algorithm}
\usepackage{hhline}
\usepackage{algpseudocode}
\usepackage{multirow}
\usepackage{array}
\usepackage{caption}
\usepackage{listings}
\usepackage{diagbox}
\usepackage{tikz}
\usepackage[colorlinks=true,linkcolor=red,citecolor=blue]{hyperref}
\usepackage[utf8]{inputenc}
\usepackage{bm}
\usetikzlibrary{spy}

\usepackage{pgfplots}
\pgfplotsset{compat=newest}
\usepgfplotslibrary{groupplots}
\usepgfplotslibrary{dateplot}
\usepackage[caption=false,labelformat=empty]{subfig}
\newcommand{\subparagraph}{}
\input
\usepackage{makecell}
\makeatletter
\def\BState{\State\hskip-\ALG@thistlm}
\makeatother
\usepackage{setspace}

\DeclarePairedDelimiter\floor{\lfloor}{\rfloor}

\newtheorem{theorem}{Theorem}

\makeatletter
\def\Cline#1#2{\@Cline#1#2\@nil}
\def\@Cline#1-#2#3\@nil{%
  \omit
  \@multicnt#1%
  \advance\@multispan\m@ne
  \ifnum\@multicnt=\@ne\@firstofone{&\omit}\fi
  \@multicnt#2%
  \advance\@multicnt-#1%
  \advance\@multispan\@ne
  \leaders\hrule\@height#3\hfill
  \cr}
\makeatother

\makeatletter
\def\BState{\State\hskip-\ALG@thistlm}
\makeatother
\newcommand{\paragraphb}[1]{\vspace{0.03in} \noindent{\bf #1} }

\usepackage{colortbl}

\newcommand{\fade}[1]{\textcolor{gray}{#1}}

\begin{document}

\title{Cronus: Robust and Heterogeneous Collaborative Learning with Black-Box Knowledge Transfer}
\author{Hongyan Chang\thanks{* Equally contributing authors. Part of the work was done when Virat Shejwalkar was a research intern at NUS.}$^{*1}$, Virat Shejwalkar$^{*2}$, Reza Shokri$^{1}$, Amir Houmansadr$^{2}$\\ $^{1}$National University of Singapore, $^{2}$University of Massachusetts Amherst\\ {$^{1}$\{hongyan, reza\}@comp.nus.edu.sg, $^{2}$\{vshejwalkar, amir\}@cs.umass.edu}}
\IEEEoverridecommandlockouts
\makeatletter\def\@IEEEpubidpullup{9\baselineskip}\makeatother

\maketitle

\begin{abstract}
Collaborative (federated) learning enables multiple parties to train a global model without sharing their private data, but through repeated sharing of the parameters of their local models.  Each party updates its local model using the aggregation of all parties' parameters, before each round of local training. 

Despite its advantages, this approach has many known privacy and security weaknesses and performance overhead, in addition to being limited only to models with homogeneous architectures.  Shared parameters leak a significant amount of information about the local (and supposedly private) datasets.  Besides, federated learning is severely vulnerable to poisoning attacks, where some participants can adversarially influence the aggregate parameters.  Large models, with high dimensional parameter vectors, are in particular highly susceptible to privacy and security attacks: {\em curse of dimensionality} in federated learning.  We argue that sharing parameters is the most naive way of information exchange in collaborative learning, as they open all the internal state of the model to inference attacks, and maximize the model's malleability by stealthy poisoning attacks.

We propose \emph{Cronus}, a robust collaborative machine learning framework.  The simple yet effective idea behind designing Cronus is to control, unify, and significantly reduce the dimensions of the exchanged information between parties, through robust knowledge transfer between their black-box local models.  We evaluate all existing federated learning algorithms against poisoning attacks, and we show that Cronus is the only secure method, due to its tight robustness guarantee.  Treating local models as black-box, reduces the information leakage through models, and enables us using existing privacy-preserving algorithms that mitigate the risk of information leakage through the model's output (predictions).  Cronus also has a significantly lower sample complexity, compared to federated learning, which does not bind its security to the number of participants.  It also allows collaboration between models with heterogeneous architectures.
\end{abstract}

\section{Introduction}

Collaborative machine learning has recently emerged as a promising approach for building machine learning models using distributed training data held by multiple parties.  The training is distributed, and participants repeatedly exchange information about their local models, through an aggregation server.  The objective is to enable all the participants to converge to a global model, while keeping their data private.  This is very attractive to parties who own sensitive data, and agree on performing a common machine learning task, yet are unwilling to pool their data together for centralized training.  Various applications can substantially benefit from collaborative learning.  Examples include medical and financial applications, intelligent virtual assistants, speech recognition, keyboard input prediction, and mobile vision \cite{mcmahan2018learning, mcmahan2017communication}.

A popular approach for collaborative deep learning, known as federated learning, assumes \emph{homogeneous} local models, i.e., with the same architecture~\cite{shokri2015privacy, mcmahan2017communication}.  Every model shares its parameters (or gradients) with a parameter server, after each rounds of training on its local data.  The server aggregates the parameter vectors by computing their element-wise mean, and shares the aggregate with the participants. Each party updates its local model with the latest global aggregate, and continues with the next round of local training.

There are major obstacles hindering the scalable deployment of secure and truly privacy-preserving federated learning for sensitive applications.  Existing federated learning algorithms are not robust to adversarial updates \cite{bhagoji2019analyzing, blanchard2017machine} and backdoor attacks \cite{bhagoji2019analyzing, bagdasaryan2018how}, and can leak a significant amount of sensitive information about local datasets \cite{nasr2019comprehensive, melis2019exploiting}.  Besides, federated learning cannot be used for aggregating heterogeneous models, for participants that use different models as they have different memory and computing power.

In this paper, our main focus is on protecting collaborative machine learning against poisoning attacks.  There exist a long chain of recent poisoning attacks and defenses for federated learning~\cite{mhamdi2018the, xie2018generalized, yin2018byzantine, blanchard2017machine, wagner2004resilient, li2014resolving, nasr2019comprehensive, bhagoji2019analyzing, bagdasaryan2018how}.  All existing defenses focus only on replacing the vulnerable  aggregation in federated learning with a robust mean function, leaving the rest of the framework intact.  But, as it is shown in the literature, and as we show in this paper, all existing (aggregation) algorithms in federated learning are susceptible to some form of poisoning attack, which can significantly damage the global model.

By sharing the model parameters, federated learning completely opens the local models to the (potentially malicious) server and untrusted participants. 
This is a serious privacy vulnerability as the model parameters leak all the information that the model has about its training data \cite{nasr2019comprehensive,melis2019exploiting}. 
The problem is significantly worse for large models, with huge number of parameters. 
This high dimensionality not only magnifies the problem, but also makes it harder to design a defense mechanism with tight (robustness and privacy) guarantees.  We refer to this as the \emph{curse of dimensionality} for privacy and security in federated learning.

The theoretical error bound and sample complexity\footnote{Sample complexity, in the context of federated learning, is the number of parties required to achieve the theoretical error bound.} of the existing robust aggregation algorithms \cite{blanchard2017machine, mhamdi2018the, diakonikolas2016robust, diakonikolas2017being, diakonikolas2019sever, li2018principled} depend on the dimensionality of model parameters.
The high dimensionality of the models makes the error bound and sample complexity of these robust aggregation algorithms prohibitively high, which makes these algorithms susceptible to some form of poisoning attacks.  Besides, federated learning algorithms require overwriting the local models' parameters by the global model.  This makes them extremely susceptible to poisoning and backdoor attacks, as the adversary can make small changes in the high-dimensional models which damage the model, but are hard to detect. 

Thus, sharing model parameters to transfer the knowledge of local models is a wrong design choice in federated learning, especially considering that this is not the only way of exchanging knowledge between models.

\textbf{\emph{Our contributions.}} In this paper, we design \emph{Cronus}, a collaborative learning approach to address the fundamental shortcomings of federated learning.  Instead of sharing parameters with the server, and updating them by overwriting them with the aggregated parameters, we extract, aggregate, and transfer the knowledge of models in a black-box manner.  We use knowledge transfer algorithms, which are used for various compression and regularization purposes in machine learning~\cite{hinton2014distilling, ba2014deep, wang2018kdgan, anil2018large}, or as part of a low-sensitivity algorithm to train a model in a centralized setting with differential privacy~\cite{papernot2017semi, papernot2018scalable}. 

In Cronus, we support {\bf heterogeneous} model architectures, as knowledge transfer through distillation allows sharing models via their predictions.  For this purpose, we assume that an unlabeled public dataset is available.  After a few rounds of local training on their private data, the Cronus parties share their predictions on the public data. 

Local models are {\em fine-tuned} using the aggregated predictions (instead of directly overwriting their parameters), which significantly reduces the potential of harmful updates on local models. Besides, Cronus reduces the dimensions of the update vectors to the size of the model's output, as opposed to the size of the model's parameters.  This reduces the dimensionality of the vectors in robustness algorithms, by many orders of magnitudes.  Thus, we are able to provide tight provable {\bf robustness} guarantees on the aggregation algorithms, even for small networks. with a dozen of participants. 

The distillation process, using disjoint data to the models' training set, is used as a regularization technique~\cite{ba2014deep}, and can reduce information leakage about the training data~\cite{shokri2017membership}.  More importantly, the black-box nature of our algorithm also allows for using privacy-preserving mechanisms by the parties that are tailored to specifically protect the {\bf privacy} of local training data in the black-box setting~\cite{nasr2018machine, dwork2018privacy}.  

Using benchmark datasets, we comprehensively compare the robustness of Cronus with existing federated learning algorithms.  We evaluate them against strongest poisoning attacks in the literature. We also design new attacks that breaks a class of aggregation algorithms (based on multiplicative weights updates \cite{arora2012the,li2014resolving}).  We show that each and every of the aggregation schemes used in federated learning (and proposed to make it robust) can be broken using some form of poisoning attacks.  The models' accuracy under attack is significantly reduced, sometimes to that of a random guess.  However, none of the poisoning attacks can impact the accuracy of models learned using Cronus leaning, beyond at most 2\% of drop in prediction accuracy.  For the strongest attack on Cronus learning, the reduction in accuracy for Purchase models is 1.6\%, for SVHN model is 1.3\%, for MNIST model is 1.5\%, and for CIFAR10 is 2.1\%.

Due to sharing knowledge through predictions, Cronus significantly reduces the risk of active and passive membership inference attacks.
The Cronus learning also allows for local party models to prevent sensitive information leakage through their predictions.  We train local models with membership privacy through adversarial regularization~\cite{nasr2018machine}.  Using this approach, we show that, for the evaluated datasets, the accuracy of membership inference attacks drops to near random guessing, with negligible accuracy loss.  We empirically show that the stat-of-the-art differential privacy algorithms for deep learning~\cite{abadi2016deep, mcmahan2018learning} do not provide good prediction accuracy, and are of lower accuracy even compared to stand-alone training.

Finally, we evaluate Cronus with heterogeneous model architectures, and show its effectiveness without compromising the final accuracy of party models.

\section{Threat Models}\label{problem_statement}

In this work, we investigate two important threats against federated learning algorithms:
(1) poisoning attacks against the \emph{robustness} of aggregation schemes used in federated learning~\cite{baruch2019a, bhagoji2019analyzing, mhamdi2018the}, and (2) inference attacks against the \emph{privacy} of models' sensitive training data~\cite{nasr2019comprehensive, melis2019exploiting}.
 
\label{ps:threat_model}

\subsection{Poisoning attacks}\label{ps:threat_model:poisoning}

The aim of the poisoning attacks is to indiscriminately jeopardize the accuracy of the resulting models of the federated learning.  To achieve this, the adversary controls an $\epsilon$ fraction of the total $n$ parties and sends malicious updates to the server through the $\epsilon n$ malicious parties.  We assume that the adversary also has access to some data which is drawn from the same distribution as that of the local training data of the $(1-\epsilon) n$ \emph{benign parties}.  Therefore, the adversary has the upper hand in that it can obtain benign updates, and use them to craft effective malicious updates using various poisoning attacks (Section~\ref{attacks}).  Each (robust) aggregation algorithm has a breaking point with respect to $\epsilon$, i.e., it cannot provide any robustness guarantee if $\epsilon$ is larger than the breaking point.  We set the fraction of the attackers to be the maximum value which is less than the breaking point, so the robust aggregation algorithms can be evaluated against the strongest attack that they claim they can protect against.

\subsection{Inference attacks}\label{ps:threat_model:meminf}

In federated learning, we consider the information leakage through each party's update, and through the aggregated updates, about the party's local data.  The attacker can run membership inference attacks on an individual party's update or on the aggregate of updates from all parties~\cite{nasr2019comprehensive}.   The adversary's objective is to  infer if a particular sample belongs to the private training data of a single party (if target update is of a single party) or of any party (if target update is the aggregate).  Following the previous works \cite{melis2019exploiting, nasr2019comprehensive}, to evaluate the risk to a party $i$ due to a strong membership inference adversary, we assume that the adversary has knowledge of 50\% of members of $\mathbf{d}_i$ and the equal number of non-members.
The adversary then trains a binary classifier on various features of the victim models in order to learn to distinguish between the members and non-members of $\mathbf{d}_i$.

\section{Aggregation algorithms}\label{aggregation}

In this section, we first describe the setting of federated learning and, various aggregation algorithms used to combine the updates of the collaborating parties and their theoretical robustness guarantees.

\subsection{Federated learning setting}\label{collaborative_learning}

Federated learning \cite{mcmahan2017communication,shokri2015privacy} enables multiple data holders to train a global model without sharing their data, and through sharing of their training gradients/parameters~\cite{blanchard2017machine,alistarh2017qsgd,mhamdi2018the,xie2018generalized,yin2018byzantine}.
For concreteness, below we describe Federated Average (FedAvg) algorithm \cite{mcmahan2017communication} and its setting, and use it in the rest of our work.

In FedAvg, multiple parties collaborate over multiple epochs to learn a global machine learning model with a classification performance superior to the models learned individually.
FedAvg assumes that there are $n$ parties with their \emph{local training datasets}, $D_i$'s,  and a \emph{central server} which aggregates the party updates and broadcasts the aggregate to all of the parties.
In the $t^{th}$ epoch of FedAvg, parties train the aggregate $\theta^{t}_a$, broadcast by the server at the end of $t^{th}$ epoch, on their local training data, $D_i$.
Parties use stochastic gradient descent for updating, i.e., $\theta^{t}_i=\theta^{t}_a-\nabla_{\theta}L(D_i;\theta^{t}_a)$, where $L(D;\theta)$ is loss of $\theta$ on data $D$.
Each party then sends the parameters of the locally updated model,  $\theta^t_i$, to the server for aggregation.
The central server collects all the $\theta^t_i$ updates and computes their aggregate $\theta^t_a = f_\textsf{\tiny{AGG}}(\theta^t_{i\in[n]})$.
Specifically, FedAvg uses the weighted average as its $f_\textsf{\tiny{AGG}}$, where the weight of the $i^{th}$ party is determined based on the size of her local training data, i.e., $w_i=\frac{|D_i|}{|D|}$; $|D|$ denotes size of dataset $D$. The weighted average is formally given by:

\begin{align}\label{agg:average}
f_\mathsf{Mean}: \quad \theta^{t+1}_a  = \sum^n_{i=1} \frac{|D_i|}{\sum^n_{j=1}|D_j|}\theta^t_i
\end{align}

The server then broadcasts the aggregate $\theta^t_a$ to all  $n$ parties.
This process repeats for $T$ epochs or until sufficient accuracy is achieved by the aggregated global model.
The procedure is described in Algorithm \ref{alg:fed_learn}.
This algorithm is not robust against poisoning attack and even a single party can destroy the global model.
Multiple robust aggregation schemes have been proposed in the literature to improve the resilience of federated learning to poisoning attacks while maintaining the accuracy of the final model \cite{mhamdi2018the,xie2018generalized,yin2018byzantine,blanchard2017machine}.
For instance, median is a more robust statistic of data compared to mean \cite{huber2011robust,wagner2004resilient,yin2018byzantine}, and therefore, $f_\mathsf{Mean}$ in FedAvg can be replaced with weighted median aggregation $f_\mathsf{Median}$, which  computes coordinate-wise weighted median of the updates.

\begin{algorithm}[t]

	\caption{Federated learning algorithm \cite{mcmahan2017communication,shokri2015privacy}}
	\label{alg:fed_learn}
	\fontsize{9}{11}\selectfont{}
	\begin{algorithmic}[1]
	\State Initialize global model $\theta^0_a$
	\For {$t\in [T]$}
		\For {$i\in [n]$}
			\Comment{\fade{Party $i$'s local update}}
			\State $\theta^{t}_i\leftarrow \theta^{t}_a - \nabla_{\theta}L(D_i;\theta^{t}_a)$
			\State Return $\theta^{t}_i$ to server
		\EndFor

		\State $\theta^{t+1}_a = f_\textsf{\tiny{AGG}}(\theta^t_{i\in [n]})$
		\Comment{\fade{Aggregation of updates at server}}
		\State Return $\theta^{t+1}_a$ to all the parties
	\EndFor
	\end{algorithmic}
\end{algorithm}

\subsection{Krum}\label{aggregation:krum}

Weighted average aggregation cannot tolerate even a single malicious party \cite{blanchard2017machine,wagner2004resilient}.
To solve this problem, Blanchard et al. \cite{blanchard2017machine} proposed $\mathsf{Krum}$ aggregation, which is based on geometric median of vectors \cite{su2017defending,mhamdi2018the}.
The intuition behind $\mathsf{Krum}$ is as follows: $\mathsf{Krum}$ assumes that party updates have a normal distribution and that the benign updates lie close to each other in the parameter space.
Hence, instead of computing the average of the updates, $\mathsf{Krum}$ selects as aggregate the update that is closest to its $(1-\epsilon)n$ neighbor updates.
The details of the aggregation are as follows.

Let $\theta_1,...,\theta_n$ be the updates received by the server.
For $i\neq j$, $i \rightarrow j$ denotes that $\theta_j$ belongs to the $(1-\epsilon)n-2$ updates closest to $\theta_i$.
Let $s(\theta_i) = \sum_{i \rightarrow j} ||\theta_i-\theta_j||^2$ be the \emph{score} of $\theta_i$.
Then, $\mathsf{Krum}$ selects the $\theta_k$ with the lowest score.
The $\mathsf{Krum}$ aggregation algorithm is formalized in \eqref{krum}.

\begin{align}\label{krum}
f_\mathsf{Krum}: \quad \theta^{t+1}_a = \underset{\theta^t_{i\in[n]}}{\text{argmin}} \sum_{i \rightarrow j} ||\theta^t_i-\theta^t_j||^2
\end{align}

\begin{table}
\fontsize{9}{11}\selectfont{}
\begin{center}
\caption{
Theoretical error rates of aggregation algorithms. $n$ is number of parties, $\epsilon$ is breaking point, i.e.,  malicious parties' fraction, $d$ ($d_p$) is updates' dimensions, and $\sigma^2$ is the variance of each of the dimensions (assume each dimension has the same variance).
Cronus, unlike other aggregations, has error rate independent of the dimension of updates.
}

\label{tab:theory_analysis}

\begin{tabular}{|c|c|c|c|c|}
\hline
& Breaking &Statistical & Computational \\
& point & error rate & cost \\

\hline $\mathsf{Mean}$ \cite{mcmahan2017communication}
& $1/n$
& Unbounded
& ${O}(nd)$ \\
\hline $\mathsf{Median}$ \cite{li2018principled}
& $1/2$
& ${O}(\sigma \epsilon \sqrt{d})$
& ${O}(nd\log n )$ \\
\hline $\mathsf{Krum}$ \cite{blanchard2017machine}
& $(n-2)/2n$ 
& ${O}(\sigma n \sqrt{d})$
& ${O}(n^2 d)$ \\
\hline $\mathsf{Bulyan}$ \cite{mhamdi2018the}
& $(n-3)/4n$ 
& ${O}(\sigma \sqrt{d})$
& ${O}(n^2 d) $ \\
\hline
Cronus
& $1/2$
& ${O}(\sigma \sqrt{\epsilon})$
& ${O}(d_p^3 + n)$ \\

\hline
\end{tabular}
\end{center}

\end{table}

\subsection{Bulyan}\label{aggregation:bulyan}

The breaking point of $\mathsf{Krum}$ is $\epsilon=(\frac{n-2}{2n})$, i.e., it can  tolerate up to $(\frac{n-2}{2})$ malicious parties, while maintaining high utility of the final model \cite{blanchard2017machine}.
However, El Mhamdi et al. \cite{mhamdi2018the} proposed an attack on $\mathsf{Krum}$ assuming an omniscient adversary who has access to all the benign updates.
The attack exploits the fact that, in a vector space of dimension $d\gg1$, small disagreements on each coordinate translate into a distance $\Vert \mathbf{x}-\mathbf{y}\Vert_p=\mathcal{O}(\sqrt[p]{d})$.
Therefore, the adversary crafts a malicious update with a single dimension set to a large value, and the other dimensions set to the average of the benign updates.
Such malicious update pushes the parameter vector to a sub–optimal parameter space and destroys the global model's accuracy.

Essentially, $\mathsf{Krum}$ filters outliers based on the entire update vector, but does not filter coordinate-wise outliers.
To address this, \cite{mhamdi2018the} proposes a meta-aggregation rule $\mathsf{Bulyan}$, which performs vector-wise, e.g. $\mathsf{Krum}$, and coordinate-wise, e.g. $\mathsf{TrimmedMean}$ \cite{yin2018byzantine}, filtering in two steps.  At first, $\mathsf{Bulyan}$ uses some Byzantine resilient aggregation $\mathcal{A}$, e.g., $\mathsf{Krum}$ in Algorithm \ref{alg:bulyan},  to filter outliers based on the distances between the update vectors, and then aggregates these updates using a variant of $\mathsf{TrimmedMean}$.
Algorithm~\ref{alg:bulyan} describes the $\mathsf{Bulyan}$ aggregation.

\begin{algorithm}
\caption{Bulyan aggregation: $f_\mathsf{Bulyan}$ \cite{mhamdi2018the}}
\label{alg:bulyan}
\fontsize{9}{11}\selectfont{}
\begin{algorithmic}[1]
\State \textbf{Input}: $\mathcal{A} = f_\mathsf{Krum}$, $\mathcal{P}=(\theta^t_1,...,\theta^t_n)$, $n$, $\epsilon$
\State $S\leftarrow \emptyset$
\While{$|S|<(1-2\epsilon)n$}
	\State $p \leftarrow \mathcal{A}(\mathcal{P}\backslash S)$
	\State $S\leftarrow S\cup \{p\}$
\EndWhile
\State \textbf{Output}: $\theta^{t+1}_a = \mathsf{\mathsf{TrimmedMean}}(S)$
\end{algorithmic}
\end{algorithm}

Among the different variants of $\mathsf{TrimmedMean}$ \cite{xie2018generalized,mhamdi2018the,yin2018byzantine}, we follow the one used in the original work \cite{mhamdi2018the} given in \eqref{trimmed_mean}.
Here, $U_j$ is defined as the set of indices of the top-$(1-2\epsilon)n$ values in $(\theta^j_1,...,\theta^j_n)$ nearest to their \emph{median} $\mu_j$.

\begin{equation}\label{trimmed_mean}
\mathsf{TrimmedMean}(\theta_1,...,\theta_N) =\bigg\{\theta^j_a=\frac{1}{|U_j|} \sum_{i\in U^j} \theta_i^j \ \forall j\in [d]  \bigg\}
\end{equation}

\subsection{Multiplicative weight update ($\mathsf{MWU}$)}\label{aggregation:mwu}

Multiplicative weight update ($\mathsf{MWU}$) technique lies at the core of many learning algorithms \cite{arora2012the,freud1997decision,plotkin1991fast,garg2007faster,li2014resolving}.
The general framework of $\mathsf{MWU}$ is given in Algorithm \ref{alg:mwu}.
The intuition behind $\mathsf{MWU}$-based aggregations is to reduce the weights of malicious parties using the distance between their malicious updates and the aggregated update.
This is based on two assumptions: malicious updates lie farther away from the mean compared with the benign updates, and the number of malicious updates is smaller than that of the benign updates.
Therefore, $\mathsf{MWU}$-based aggregation schemes have the breaking point of $\floor{\frac{n-1}{2}}$ malicious parties.

\begin{algorithm}

\caption{Multiplicative weights update: $f_\mathsf{MWU}$}

\label{alg:mwu}

\fontsize{9}{11}\selectfont{}
\begin{algorithmic}[1]
\State \textbf{Input}: $\mathcal{P}=(\theta^t_1,...,\theta^t_n)$
\State Initialize  parties' weight vector $\mathbf{w}^0\leftarrow \mathbf{1}$ and $\theta_a^{0}\leftarrow f_\mathsf{AGG}( \mathbf{w}^0,\mathcal{P})$ at $t=0$

\Repeat

\State $\mathbf{w}^{t+1}\leftarrow$ $\mathsf{WeightUpdate}$ ($\mathbf{w}^t$, $\theta_a^{t},\mathcal{P}$)

\State $\theta_a^{t+1} \leftarrow f_\mathsf{AGG}( \mathbf{w}^{t+1},\mathcal{P})$

\State $t \leftarrow t+1$

\Until {Convergence criterion is satisfied}

\State \textbf{Output}: final $\theta_a$
\end{algorithmic}
\end{algorithm}

There are different variants of $\mathsf{MWU}$ that use different functions for $\mathsf{WeightUpdate}$ and $f_\mathsf{AGG}$ in Algorithm \ref{alg:mwu}.
We detail two of them next.

\subsubsection{$\mathsf{MWU}$ with mean aggregation}\label{agg:mwu_avg}

In $\mathsf{MWU}$, if the weighted mean (Section \ref{collaborative_learning}) is used as the aggregation algorithm, $f_\mathsf{AGG}$ in Algorithm \ref{alg:mwu}, it is called $\mathsf{MwuAvg}$.
Here, the weight of the $i^{th}$ party is updated based on the distance between the weighted average, $\theta_a^t$, of all the updates and $\theta_i$; this is given by \eqref{mwu_avg_wu}.
The weights of all the parties are equal at the beginning.

\begin{align}
\label{mwu_avg_wu}
w^{t+1}_i &= w^{t}_i \ \text{exp}(- ||\theta_a^t-\theta_i||_p)\\
\label{mwu_avg_agg}
\theta_a^{t+1} &= \frac{\sum^n_{i=1}w_i^{t+1} \theta_i}{\sum^n_{i=1}w_i^{t+1}}
\end{align}

\subsubsection{$\mathsf{MWU}$ with optimization}\label{agg:mwu_opt}

Li et al. \cite{li2014resolving} propose a truth discovery  framework CRH, to aggregate the responses in a crowd-sourcing setting.
In each epoch, the framework updates the weights of parties based on the solution of an optimization problem.
Essentially, the weight update algorithm considers the distance of parties' updates from the aggregate of all the updates in each epoch.
The weight update and aggregate computation are given by \eqref{mwu_opt_wu} and \eqref{mwu_opt_agg}, respectively.
For further details of the aggregation, please refer to \cite{li2014resolving}.

\begin{align}
\label{mwu_opt_wu}
w^{t+1}_i &=-\text{log}\Bigg(\frac{||\theta^t-\theta_i||_p}{\sum^n_{i=1}||\theta^t-\theta_i||_p}\Bigg)\\
\label{mwu_opt_agg}
\theta^{t+1} &= \sum^n_{i=1}w_i^{t+1} \theta_i
\end{align}

\section{Attacks on Aggregation Algorithms}\label{attacks}

In this section, we detail the poisoning and membership inference attacks used in our work to evaluate the robustness and privacy in federated learning.
The poisoning attacks are of two types: \emph{availability} and \emph{targeted} attacks.
The earlier attacks aim to jeopardize the overall accuracy of the final model/s \cite{baruch2019a,hayes2018contamination,steinhardt2017certified}, while the latter attacks aim to mis-classify only a specific set of samples of the attacker's choice at the test time \cite{bagdasaryan2018how,bhagoji2019analyzing}.
We focus on the poisoning availability attacks and below introduce such attacks from the literature, and also introduce a new poisoning attack targeting the $\mathsf{MwuAvg}$ and $\mathsf{MwuOpt}$ aggregations.

\subsection{Label flip poisoning (Label flip)}\label{attack:lf_poison}

We consider a type of data poisoning attacks \cite{jagielski2018manipulating,munoz2017towards,hayes2018contamination}, where the adversary flips the labels of her local training data in a particular fashion to poison it.
We call this attack \emph{label flip poisoning} attack.
The label flipping strategy is performed consistently across all of the $\epsilon n$ malicious parties that the adversary controls, i.e., all the malicious parties flip labels in the exact same way.
Then, the $\epsilon n$ malicious parties use this poisoned data to train their local models and then share corresponding updates with the central server.

\begin{algorithm}[b]
\caption{Little is enough attack (LIE) \cite{baruch2019a}}
\label{alg:lie}
\fontsize{9}{11}\selectfont{}
\begin{algorithmic}[1]
\State \textbf{Input}: $n$, $\epsilon$, mean and variance vectors of benign updates $\bar{\mu}$, $\bar{\sigma}$
\State Number of workers required for majority:
\vspace{-.3em}
$$s=\floor{\frac{n}{2}+1}-\epsilon n$$
\vspace{-.6em}
\State Using $z$-table, set $z=\underset{z}{max}\Big(\phi(z)<\frac{n-s}{n}\Big)$
\For {$j \in [d]$}
	\State $\theta_m^j\leftarrow \bar{\mu}_j+z\bar{\sigma}_j$
\EndFor
\State \textbf{Output}: malicious update $\theta_m$
\end{algorithmic}
\end{algorithm}

\subsection{Naive poisoning (PAF)}\label{attack:naive_poison}

Our threat model from Section \ref{ps:threat_model} considers an omniscient adversary who knows the distribution of  benign updates, i.e., the mean and standard deviation of each dimension of benign updates.
The adversary can estimate this distribution as she possesses data drawn from the distribution same as that of benign parties.
Given this, the adversary crafts the malicious update $\theta_m$ to be arbitrarily far from the mean of the benign updates:

\begin{equation}\label{naive_poison}
\theta_m = \frac{\sum^n_{i=1} \theta_i}{(1-\epsilon)n} + \theta'
\end{equation}

This malicious update, $\theta_m$, is then shared by each malicious party with the central server.
In \eqref{naive_poison}, $\theta'$ is a vector of size $|\theta_i|$ with arbitrarily large coordinate values.
This attack can jeopardize the weighted averaging, and interestingly, weighted median aggregations based federated learning (Section \ref{collaborative_learning}).

\subsection{Little is enough attack (LIE)}\label{attack:lie}

Baruch et al. \cite{baruch2019a} propose an attack called \emph{little is enough} (LIE).
The attack successfully circumvents state-of-the-art robust aggregation algorithms, including $\mathsf{Bulyan}$ and $\mathsf{Krum}$.
These aggregations are vulnerable to the attack, because they are tailored to an adversary that crafts a malicious update with  at least one arbitrarily large dimension.
However, in practice, a malicious update, $\theta^m$, obtained by small perturbations in a large number of dimensions of a benign update suffice to affect model's convergence and also circumvent the defenses.
Therefore, note that, \emph{the root cause of the success of the LIE attack is also the high dimensionality of the updates.}
The attack is described in Algorithm. \ref{alg:lie}.
El Mhamdi et al. \cite{mhamdi2018the}  propose an attack against geometric median based aggregations such as $\mathsf{Krum}$, but does not work against $\mathsf{Bulyan}$, and therefore is omitted from the evaluation.

\subsection{Our poisoning attack (OFOM)}\label{attack:our_attack}

In this section, we propose an attack which targets aggregation schemes that perform weighted aggregation of data by assigning the weights based on the distance of the data points from an aggregate of the data.
These aggregations are robust to all the above attacks, as we show in our evaluation. 
We discussed two such aggregation schemes: MWU with averaging \cite{arora2012the} and MWU with optimization \cite{li2014resolving} in Section \ref{aggregation}. 
In any given epoch, the aforementioned aggregation schemes start with equal weights to all parties and update a party's weight based on the distance of the party's update from weighted average of all party updates.
\emph{The attack exploits the fact that, all parties are given equal weights to start with.}
The OFOM attack craft two malicious updates:
The first update, $\theta^m_1$, is arbitrarily far away from the true mean, and is obtained by adding an arbitrarily large vector $\theta'$ to the mean of benign updates.
The second malicious update, $\theta^m_2$, is at the empirical mean of benign updates and $\theta^m_1$.
The malicious updates are formalized in \eqref{at_on_avg}.

\begin{align}\label{at_on_avg}
\theta^m_1 = \frac{\sum^n_{i=1} \theta_i}{n} + \theta', \quad \theta^m_2 = \frac{\sum^n_{i=1} \theta_i+\theta^m_1}{n+1}
\end{align}

This way, at the end of the first epoch of the MWU aggregation, the adversary manages to assign a weight close to 1 to the parties with update $\theta^m_2$.
In the case of $\mathsf{MWUAvg}$ and $\mathsf{MWUOpt}$, all the benign parties are assigned negligible weights, which completely jeopardizes the accuracy of aggregation.
To be effective, the adversary needs just two malicious parties who share the two malicious updates.

\subsection{Membership inference attacks} \label{attacks:grad_ascent}

Recent research has shown the susceptibility of the federated learning to active and passive inference attacks \cite{melis2019exploiting,nasr2019comprehensive}.
In the passive case, the attacker, either the server or some of the parties, simply observes the updated model parameters to mount membership inference attacks.
In the active case, however, the attacker tampers with the training of the victim model/s in order to  infer membership of target data in \emph{any of the benign party's data}.
Specifically, the attacker shares malicious updates and forces the victim model/s to share more information about the members of their training data that are of the attacker's interest.
This attack, called \emph{gradient ascent} attack \cite{nasr2019comprehensive}, exploits that the SGD optimization  updates model parameters in the opposite direction of the  gradient of the  loss over the private training data.
Let $\mathbf{x}$  be a record of attacker's interest and $\theta_a$ be the current global model.
The attacker \emph{crafts the malicious update $\theta^m$ by updating parameters of $\theta_a$ in the same direction of the gradient of the loss on $\mathbf{x}$}, i.e., performs gradient ascent as: $\theta^m =\theta_a + \gamma \frac{\partial L_\mathbf{x}}{\partial \theta_a}$.
Such $\theta^m$, when combined with the benign updates, increases the loss of the resulting global model, $\theta'_a$, on $\mathbf{x}$.
If $\mathbf{x}$ is in the training data of some party, SGD on $\theta'_a$ by this party will sharply reduce the loss of $\mathbf{x}$.
On the other hand, if $\mathbf{x}$ is not in any party's training data, the loss of $\mathbf{x}$ will remain almost unchanged. Therefore, this attack increases the gap between the losses of $\theta_a$ on members and non-members and facilitates membership inference.

\section{Cronus}\label{cronus}

In the existing federated learning algorithms, described in Section \ref{collaborative_learning}, the server repeatedly collects the parameters of the local models, aggregates them by computing their mean, and sends the aggregate parameter vector back to the parties.  This is the simplest way of sharing knowledge between the participants, and has the following fundamental drawbacks:

{\bf Robustness:} As shown in Table~\ref{tab:theory_analysis}, the upper bound on the error of aggregation algorithms (in the adversarial setting) depends on the dimensionality of the model parameters. This makes the aggregated models to be highly error-prone, for large models. Thus, they are very susceptible to poisoning attacks, as described in Section~\ref{attacks}.

{\bf Privacy:} Sharing the model parameters facilitates strong white-box inference attacks, as it opens up the model to the adversary~\cite{melis2019exploiting, nasr2019comprehensive}. The larger the models are, the more significant their leakage about their local training data is.

{\bf Heterogeneity:} Parameter aggregation is restricted to homogeneous architectures, i.e., all parties need to have the same model architecture.

Instead of sharing the raw models at each round of training, the participants would need to share the knowledge that they have learned from their training data, in a succinct way.  This is the main objective of \emph{knowledge transfer} \cite{hinton2014distilling,ba2014deep,papernot2017semi,papernot2018scalable,wang2018kdgan,anil2018large}, which is a powerful tool to share knowledge of a model through its predictions.

Knowledge transfer efficiently transfers the represented function by a model (or an ensemble of models) to a student model \cite{hinton2014distilling}.  It makes learning very effective for the student model by placing equal weight on the relationships learned by the teachers across all the classes and significantly improves the convergence of student model (as compared to training directly on hard-labeled data)~\cite{hinton2014distilling, ba2014deep, anil2018large}.  Furthermore, knowledge transfer is an effective regularization method~\cite{hinton2014distilling}.

What makes this approach, in particular, suitable for {\bf robust,  privacy-preserving, and heterogeneous} federated learning is three-fold. First, it significantly reduces the dimensions of the updates, from the size of the model parameters, to its output size. This enables using robust aggregation algorithms with very tight robustness guarantee. Second, it limits the interactions with each local model to the black-box access setting. Thus, the negative impact of poisoning attacks {\em onto} the model, and the information leakage {\em from} its training data, are both limited.  The knowledge transfer through the predictions on a dataset, that does not overlap with the model's training set, itself reduces the information leakage of a model about its training data~\cite{shokri2017membership}.  In addition, the black-box setting allows us to make use of utility-preserving algorithms to make the knowledge transfer privacy-preserving, as a party only needs to mitigate the privacy risk of sharing its predictions~\cite{nasr2018machine, papernot2017semi, dwork2018privacy}.  Third, all the models agree on a particular learning task, so are homogeneous on their output vectors.  Yet, they can be of heterogeneous architectures, or even different families of machine learning algorithms.

In this paper, we leverage knowledge transfer and propose the {\bf\emph{Cronus} federated learning} algorithm, where the aggregation algorithm devours the knowledge of local models, and disgorges their robust aggregation.  The parties update their models using the aggregated knowledge as well as their local data. Then, each party shares their knowledge through (privacy-preserving) predictions of their local models on a public dataset.  This repeats in every round of the federated learning.

\subsection{Cronus Collaborative Learning}\label{cronus:protocol}

In addition to the local private datasets of parties, Cronus assumes a set of \emph{unlabeled} public data, $X_p$.  Such public data is essentially a set of feature vectors that Cronus uses to extract and exchange the knowledge of local models.

Algorithm~\ref{alg:cronus} describes the Cronus collaborative learning algorithm.  Cronus has two training phases: In the first phase, called the \emph{initialization phase}, every party $i$ updates its local model parameters $\theta_i$ on its local training data $D_i$ for $T_1$ times without any collaboration.  In the second phase, called the \emph{collaboration phase}, the parties share the knowledge of their local models via their predictions on the public dataset, $X_p$.  Specifically, each epoch of this phase includes:

\begin{itemize}
\item Each party computes soft labels for $X_p$, using its local model parameters $\theta_i$ to get prediction vectors $Y_i$ and shares them with the server.
\item The server aggregates the predictions (separately for each public data), i.e., computes $\bar{Y} =f_\mathsf{Cronus}(Y_i,..,Y_n)$, and sends  $\bar{Y}$ to all parties; $f_\mathsf{Cronus}$ is an aggregation algorithm, which we describe in Section~\ref{cronus:mean_estimation}.
\item Each party updates its local model parameters $\theta_i$ using their private data $\mathbf{D}_i$ and the soft-labeled public data $(X_p,\bar{Y})$.
\end{itemize}

Sharing predictions instead of parameters completely eliminates the risk of  white-box inference attacks, as no party releases its model's parameters.  While training on their local sensitive data, parties can anticipate the privacy risks of information leakage through their predictions and \textsc{\textsf{\small{Train}}} their models with, for example, a membership privacy mechanism~\cite{nasr2018machine}.  They can also make use of prediction privacy mechanisms before sharing their predictions with the server~\cite{dwork2018privacy}.

\begin{algorithm}[t]
\caption{Cronus algorithm. Initialization phase does not involve collaboration. $D_i$ and $\theta_i$ are local dataset and model paramters from $i$-th party. $Y^t_i$ are predictions from $i$-th party on public dataset $X_p$ in epoch $t$ and $Y^t_i[k]$ is the prediction on $k$-th public data in $D_p$.}
\label{alg:cronus}
\fontsize{9}{11}\selectfont{}
\begin{algorithmic}[1]
\State \textbf{\hskip 8em Initialization phase}
\vspace{.3em}
\State Each party $i \in [n]$ updates parameters in parallel:
	\For{$t\in[T_1]$ epochs} \Comment{\fade{Training  \emph{without} collaboration}}
	\State Update $\theta_i \leftarrow$
	 \textsc{\textsf{\small{Train}}} $(\theta_i, D_i)$
	\EndFor
	\State $Y^0_{i} = \textsc{\textsf{\small{Predict}}}(\theta_i;X_p)$
	\Comment{\fade{Compute initial predictions on $X_p$}}

	\State Send $Y^0_{i}$ to the server
\vspace{1em}
\State\textbf{\hskip 8em Collaboration phase}
\vspace{.3em}
\State $\bar{Y}^0 = f_\mathsf{Cronus}(\{Y^0_{i\in[n]}\})$
\Comment{\fade{Initial aggregation at the server}}

\For{$t\in[T_2]$ epochs}

	\For{$i\in[n]$ parties} \Comment {\fade{Each party updates parameter in parallel}}

		\State $D_p = \{X_p,\bar{Y}^t\}$

		\State $\theta_i \leftarrow$ \textsc{\textsf{\small{Train}}} $(\theta_i,D_i \cup D_p)$
		\Comment {\fade{Update local model parameter $\theta_i$}}

		\State $Y^t_{i} = \textsc{\textsf{\small{Predict}}}(\theta_i;X_p)$

		\State Send $Y^t_{i}$ to the server

	\EndFor
\State $\bar{Y}^{t+1} = f_\mathsf{Cronus}(\{Y^t_{i\in[n]}\})$
\Comment{\fade{Aggregation at the server}}
\EndFor
\end{algorithmic}
\end{algorithm}

\begin{table}[t]
\vspace{-1em}
\fontsize{8}{11}\selectfont{}
\begin{center}
\setlength{\extrarowheight}{-0.01cm}
\caption{
Sample complexity, $\Theta((d/\epsilon)\log d)$, of Cronus aggregation, using~\cite{diakonikolas2017being}, for parameters and predictions updates. For parameters, $d$ is size of model; for predictions, $d$ is the number of classes in the classification task. The ratio shows that Cronus learning can achieve the same error guarantee as in federated learning, but with a network which is $5$ orders of magnitude smaller, for benchmark ML tasks.
}
\label{tab:complexities}
\begin{tabular}{ | c | c | c | c |}
\hline
Dataset & Sample complexity ratio of Federated learning over Cronus \\ \hline
SVHN & $1.2 \times 10^{5}$ \\
MNIST & $3.3 \times 10^{5}$\\
Purchase & $2.75 \times 10^{5}$ \\
CIFAR10 & $10.4 \times 10^{5}$ \\

\hline
\end{tabular}
\end{center}
\vspace{-1em}
\end{table}

It's important to note that, unlike the existing federated learning algorithms, based on FedAvg~\cite{mcmahan2017communication}, \emph{Cronus does not force a single global model} onto local models.  Instead, each local model is updated separately by improving their classification accuracy and resilience to inference attacks.  This, can further improve the robustness of local models against active attacks (such as poisoning and active inference attacks), as they do not blindly overwrite their local parameters with the aggregated knowledge.

The size of the updates and the size of the network determine the error bound for robust mean estimation algorithms.  By reducing the dimensionality of the updates, Cronus significantly reduces the guaranteed error, using any robust aggregation algorithm.  Then, the sample complexity of the aggregation algorithm determines how many participants are needed to achieve a tight error bound.  Cronus uses the aggregation algorithm proposed by Diakonikolas et al. \cite{diakonikolas2017being}, whose sample complexity is $\Theta(d\log d)$, and has the least dependence on the dimensionality of updates among existing robust mean estimation algorithms.  Table~\ref{tab:complexities} shows the sample complexities for training different ML benchmark models with federated learning versus Cronus, using Diakonikolas's algorithm.  We note that Cronus significantly reduces sample complexity (by an order of $10^5$), and therefore, unlike any existing federated learning, can achieve strong theoretical error guarantees even with a small number of parties in the network.

The computational complexity of the state-of-the-art robust aggregation algorithm is $O(d^3+n)$, where $d$ is the dimensionality of the updates~\cite{diakonikolas2017being}.  This makes it impractical to use such optimal algorithms for the high dimensional parameter updates. However, the complexity cost is negligible on Cronus, due to the small number of dimensions on its updates.

\subsection{Robust Mean Estimation in Cronus}\label{cronus:mean_estimation}
We use the robust mean estimation algorithm ~\ref{alg:dkklms_smm} proposed by Diakonikolas et al. \cite{diakonikolas2017being} in Cronus. Algorithm~\ref{alg:dkklms_smm} achieves dimension independent error guarantee and nearly optimal sample complexity as shown in the Theorem \ref{thm:dkklms_mean}.

The intuition behind Algorithm \ref{alg:dkklms_smm} (step 4-15) is that when the empirical mean of the data is corrupted, then along the corrupted direction, the empirical variance is much larger than the population variance. The algorithm finds the direction $v^*$, which has the largest variance and projects the deviation of all the inputs from the empirical mean in this direction. Then filter out a randomized fraction of the data which are farthest from the mean, $\bar{Y}_k$, along this direction. Repeat the process until the variance is not large in every direction and then output the sample mean on the subsets. We apply the robust mean estimation for each data in the public dataset $X_p$ separately.

\begin{theorem}
\label{thm:dkklms_mean}
[Theorem A.16 \cite{diakonikolas2017being}]
Let $P$ be a distribution on  $\mathbb{R}^d$ with an unknown mean vector $\mu^p$ and an unknown covariance matrix $\Sigma_p \preceq \sigma^2 I$.
Let S be an $\epsilon$-corrupted set  of samples from $P$ of size $\Theta((d/\epsilon)\log d)$.
Then, there exists an efficient algorithm that, on input S and $\epsilon >0$ , with probability 9/10 outputs $\hat{\mu}$ with $||\hat{\mu} - \mu^P||_2= O(\sigma\sqrt{\epsilon})$.
\end{theorem}

\begin{algorithm}[t]
\caption{Cronus aggregation: $f_\mathsf{Cronus}$ [Algorithm 3 \cite{diakonikolas2017being}]}
\label{alg:dkklms_smm}
\fontsize{9}{11}\selectfont{}
\begin{algorithmic}[1]
\State \textbf{Input}: $S=({Y}_1,...,{Y}_n)$, $\epsilon$, k=0
\While{$k < |X_p|$} \Comment {\fade{Compute prediction for each public data}}
\State $S_k = ({Y}_1[k],...,{Y}_n[k])$
\While{True}: \Comment{\fade{Robust mean aggregation algorithm}}
\State Compute $\bar{Y}_k$,$\Sigma_k$, the mean and covariance matrix of $S_k$.
\State Find the eigenvector $v^*$ with highest eigenvalue $\lambda^*$ of $\Sigma_k$
\If{ $\lambda^* \leq 9$}
    \State Let $\bar{Y}[k]$ =  $\bar{Y}_k$ and k = k+1
		\State break
\Else
\State Draw Z from the distribution on [0,1] with probability density function $2x$
\State Let T = $Z\max\{|v^* \cdot (Y - \bar{Y}_k)|: Y \in S_k\}$.
\State Set $S_k = \{Y\in S_k:|v^* \cdot ( Y - \bar{Y}_k)|< T\}$
\EndIf
\EndWhile
\State \textbf{Output:} $\bar{Y} = \{\bar{Y}[0],\bar{Y}[1],...,\bar{Y}[k]\}$
\EndWhile
\end{algorithmic}
\end{algorithm}
For the theoretical analysis, Algorithm \ref{alg:dkklms_smm} uses randomized filtering (step 12) and repeats until the stop condition is satisfied (step 7). The follow-up works from the same authors \cite{li2018principled,diakonikolas2019sever} suggest a simpler algorithm to obtain a better performance in practice: (1) in each iteration, remove a deterministic fraction of the data instead of a random fraction. (2) repeat the filter for constant iterations in total. In the evaluation, we filter out $\epsilon/2$ fraction of the inputs in each iteration (step 12) and repeat the filter process $2$ times (step 7) and to obtain a good performance.

\section{Experimental Setup} \label{setup}

Table \ref{tab:exp_setup} presents the datasets and corresponding model architectures, number of parties, and size of the data sets that we use in our experiments.  We defer the reader to Appendix~\ref{appdx:setup}, for the descriptions of the datasets, and details of the model architectures, and training hyper-parameters.  We use PyTorch \cite{pytorch} for our evaluations.  We evaluate the security of federated learning algorithms with respect to the following measures. 

\begin{table}[t]
\fontsize{8}{11}\selectfont{}
\begin{center}
\setlength{\extrarowheight}{-0.03cm}
\caption{Experimental setup}
\label{tab:exp_setup}

\begin{tabular}{ | c | c | c | c | c | }
\hline
\multirow{2}{*}{Dataset} & \# of benign & Model & Private data & Public \\ 
& parties & architecture & per party & data  \\ \hline

SVHN \cite{netzer2011reading} & 32 & CNN & 5,000  & 10,000 \\ 
MNIST \cite{lecun1998gradient} & 28 & FC &2,000 &  10,000\\ 
Purchase \cite{shokri2017membership} & 16 & FC & 10,000 & 10,000 \\  
CIFAR10 \cite{krizhevsky2009learning} & 16 & DenseNet & 2,500 &10,000 \\ 
\hline
\end{tabular}
\end{center}
\vspace{-1em}
\end{table}

\label{setup:metrics}

\paragraphb{Robustness.} We measure the robustness of an algorithm as the ratio between the accuracy of the model against the strongest poisoning attack, over the accuracy of the model in the benign setting.  For finding the strongest attack, we evaluate an algorithm against all the attacks which are presented in Section~\ref{attacks}.

\paragraphb{Membership inference risk.}
We measure the membership inference risk as the accuracy of the inference attack, which is the percentage of data records for which the attack model correctly predicts their membership~\cite{shokri2017membership}.   We test the attack with the same number of members and non-members. Hence, 50\% accuracy corresponds to a random guess.

\begin{table}[t]
\fontsize{8}{11}\selectfont{}
\begin{center}
\caption{Comparison of the classification accuracy of Cronus models (on average) with that of the models trained in stand-alone (on average), centralize, and FedAvg settings, in the \textbf{benign} setting.}
\label{tab:stand_alone}
\begin{tabular}{| c | c | c | c | c |}

\hline
Dataset & Stand-alone & Centralized & FedAvg & Cronus \\ \hline

SVHN & 87.5 & 96.4 & 95.9 & 91.1 \\ 
MNIST & 92.8 & 97.9 & 96.7 & 95.2\\ 
Purchase & 76.3 & 94.3 & 93.3 & 89.6 \\ 
CIFAR10 & 66.8 & 90.2 & 88.4 & 80.1 \\
\hline

\end{tabular}
\end{center}
\end{table}

\begin{table}[t]
\fontsize{8}{11}\selectfont{}
\caption{The number of malicious parties used for robustness assessment of different aggregations, based on their breaking points. The number of benign parties are shown on top of each column.}
\label{tab:num_parties}

\begin{center}
\begin{tabular}{| c | c | c | c | c |}
\hline
\multirow{3}{*}{$f_\textsf{\tiny{AGG}}$} & & \multicolumn{3}{c|}{Number of malicious parties} \\ \cline{3-5}
 & Breaking &  {SVHN} & {MNIST} & {Purchase/CIFAR10} \\  
& point & \cellcolor[gray]{0.9}32 benign & \cellcolor[gray]{0.9}28 benign & \cellcolor[gray]{0.9}16 benign \\\hline 

$\mathsf{Mean}$ & $1/n$ & 1   &1  & 1   \\ 
$\mathsf{Median}$ & $1/2$ & 31 & 27 & 15 \\ 
$\mathsf{MWU}$ & $1/2$ & 31 & 27 & 15 \\ 
$\mathsf{Krum}$ & $(n-2)/2n$ & 29 & 25 & 13 \\ 
$\mathsf{Bulyan}$  & $(n-3)/4n$ & 9 & 8 & 4 \\ 
$\mathsf{Cronus}$ & $1/2$ & 31 & 27 & 15 \\ \hline

\end{tabular}
\end{center}
\end{table}

\begin{table*}
\fontsize{8}{11}\selectfont{}
\begin{center}
\setlength{\extrarowheight}{0.01cm}

\caption{
Robustness of the federated learning using different aggregation algorithms versus Cronus. Robustness is measured as the ratio of the accuracy of the final models against the strongest attack (Section~\ref{attacks}) over the models' accuracy in the benign setting. The number of benign and adversarial parties are provided in Table~\ref{tab:num_parties}. The best accuracy for each setting (row) is highlighted. 
}

\label{tab:fedlearn}

\begin{tabular}{ c | l |c |c |c |c |c |c |c |}
\cline{2-9}

\multirow{2}{*}{Dataset} &  & \multicolumn{6}{c|}{Federated learning with various aggregation algorithms} & \multirow{2}{*}{$\mathsf{Cronus}$} \\ \cline{3-8}

&  & $\mathsf{Mean}$ & $\mathsf{Median}$ & $\mathsf{MwuAvg}$ & $\mathsf{MwuOpt}$ & $\mathsf{Bulyan}$ & $\mathsf{Krum}$ &  \\ \hline
\hline

\multirow{4}{*}{SVHN} & Accuracy (Benign) & \cellcolor[gray]{0.9}\textbf{95.9} & \cellcolor[gray]{0.9}94.8 & \cellcolor[gray]{0.9}93.9 & \cellcolor[gray]{0.9}94.4 &  \cellcolor[gray]{0.9}94.5 & \cellcolor[gray]{0.9}89.6 & \cellcolor[gray]{0.9}91.1\\ 
 & Worst accuracy (Adversarial)& \cellcolor[gray]{0.8}0.9& \cellcolor[gray]{0.8}14.5 & \cellcolor[gray]{0.8}0.9 & \cellcolor[gray]{0.8}0.7 & \cellcolor[gray]{0.8}15.5 & \cellcolor[gray]{0.8}16.2 & \cellcolor[gray]{0.8}\textbf{89.8} \\ 
 & Strongest attack & (OFOM) & (LIE) & (OFOM) & (OFOM) & (LIE) & (LIE) & (Label flip)\\
 & Robustness & \cellcolor[gray]{0.95}0.01 & \cellcolor[gray]{0.95}0.15 & \cellcolor[gray]{0.95}0.01 & \cellcolor[gray]{0.95}0.01 & \cellcolor[gray]{0.95}0.16 & \cellcolor[gray]{0.95}0.18 & \cellcolor[gray]{0.95}\textbf{0.99} \\
\hline
\hline

\multirow{4}{*}{MNIST} & Accuracy (Benign) & \cellcolor[gray]{0.9}96.7  & \cellcolor[gray]{0.9}96.5 & \cellcolor[gray]{0.9}\textbf{97.2} & \cellcolor[gray]{0.9}97.4 & \cellcolor[gray]{0.9}96.9 & \cellcolor[gray]{0.9}93.3 & \cellcolor[gray]{0.9}95.2\\ 
 & Worst accuracy (Adversarial) & \cellcolor[gray]{0.8}9.6 & \cellcolor[gray]{0.8}91.5 & \cellcolor[gray]{0.8}25.3 & \cellcolor[gray]{0.8}12.7 & \cellcolor[gray]{0.8}\textbf{94.1} & \cellcolor[gray]{0.8}89.9 & \cellcolor[gray]{0.8}93.7 \\ 
 & Strongest attack & (PAF)  & (PAF)  & (OFOM)  & (PAF)  & (LIE) & (Label flip) & (Label flip) \\
 & Robustness & \cellcolor[gray]{0.95}0.09 & \cellcolor[gray]{0.95}0.95 & \cellcolor[gray]{0.95}0.26 & \cellcolor[gray]{0.95}0.13 & \cellcolor[gray]{0.95}0.97 & \cellcolor[gray]{0.95}0.96 & \cellcolor[gray]{0.95}\textbf{0.99} \\
\hline
\hline

\multirow{4}{*}{Purchase} & Accuracy (Benign) & \cellcolor[gray]{0.9}93.3  & \cellcolor[gray]{0.9}93.0 & \cellcolor[gray]{0.9}\textbf{93.6} & \cellcolor[gray]{0.9}92.5 & \cellcolor[gray]{0.9}92.8 & \cellcolor[gray]{0.9}72.1 & \cellcolor[gray]{0.9}89.6 \\ 
 & Worst accuracy (Adversarial) & \cellcolor[gray]{0.8}1.1 & \cellcolor[gray]{0.8}12.5 & \cellcolor[gray]{0.8}1.8 & \cellcolor[gray]{0.8}1.1 & \cellcolor[gray]{0.8}81.8 & \cellcolor[gray]{0.8}49.6 & \cellcolor[gray]{0.8}\textbf{88.0}\\ 
 & Strongest attack & (PAF)  & (PAF)  & (OFOM)  & (OFOM)  & (LIE) & (LIE) & (Label flip) \\
 & Robustness & \cellcolor[gray]{0.95}0.01 & \cellcolor[gray]{0.95}0.75 & \cellcolor[gray]{0.95}0.02 & \cellcolor[gray]{0.95}0.01 & \cellcolor[gray]{0.95}0.87 & \cellcolor[gray]{0.95}0.69 & \cellcolor[gray]{0.95}\textbf{0.98} \\
\hline
\hline

\multirow{4}{*}{CIFAR10} & Accuracy (Benign) & \cellcolor[gray]{0.9}88.4 & \cellcolor[gray]{0.9}\textbf{89.1} & \cellcolor[gray]{0.9}86.2 & \cellcolor[gray]{0.9}87.6 &  \cellcolor[gray]{0.9}89.0 & \cellcolor[gray]{0.9}84.5 & \cellcolor[gray]{0.9}80.1 \\ 

 & Worst accuracy (Adversarial) & \cellcolor[gray]{0.8} 11.3 & \cellcolor[gray]{0.8}15.1 & \cellcolor[gray]{0.8}14.2 & \cellcolor[gray]{0.8}12.8 & \cellcolor[gray]{0.8}75.6 & \cellcolor[gray]{0.8}18.0 & \cellcolor[gray]{0.8}\textbf{78.0} \\ 
& Strongest attack & (PAF) & (PAF) & (OFOM) & (OFOM) & (LIE) & (LIE) & (LIE)\\

& Robustness & \cellcolor[gray]{0.95}0.13 & \cellcolor[gray]{0.95}0.17 & \cellcolor[gray]{0.95}0.16 & \cellcolor[gray]{0.95}0.15 & \cellcolor[gray]{0.95}0.85 & \cellcolor[gray]{0.95}0.21 & \cellcolor[gray]{0.95}\textbf{0.97} \\
\hline
\end{tabular}
\end{center}
\end{table*}

\section{Empirical results}\label{exp}
\label{exp:cronus_performance}

In this section, we first present the performance of the models in the benign setting, as the baseline.  Then, we compare Cronus with other algorithms, with respect to their robustness, privacy, heterogeneity, and communication efficiency.  
 
Table \ref{tab:stand_alone} shows the classification accuracy of the models trained using Cronus, versus stand-alone, centralized, and FedAvg federated learning, all in the benign setting.  In the stand-alone setting, each party trains its only on its local data, without any collaboration.  In the centralized learning, a single model is trained on the union of the participants' data.  FedAvg federated learning is described in Section \ref{problem_statement}.  For stand-alone setting, and also for Cronus, the accuracy of models for different parties is different.  In these two cases, we report the average classification accuracy of the models.

\subsection{Robustness}\label{exp:robustness}

We compare the robustness of federated learning, using different aggregation algorithms (Section~\ref{aggregation}), against an adversary who mounts the strong poisoning attacks (described in Section~\ref{attacks}). To assess the worst case robustness, we evaluate algorithms against the largest fraction of malicious parties which is smaller than the breaking point.  The number of benign and malicious parties are given in Table \ref{tab:num_parties}.

The robustness assessment results are shown in Table \ref{tab:fedlearn}.  Each column corresponds to one aggregation algorithm, and the \emph{Worst accuracy} row shows the accuracy of the final models when the attack in the \emph{Strongest attack} row is mounted.

\begin{figure*}
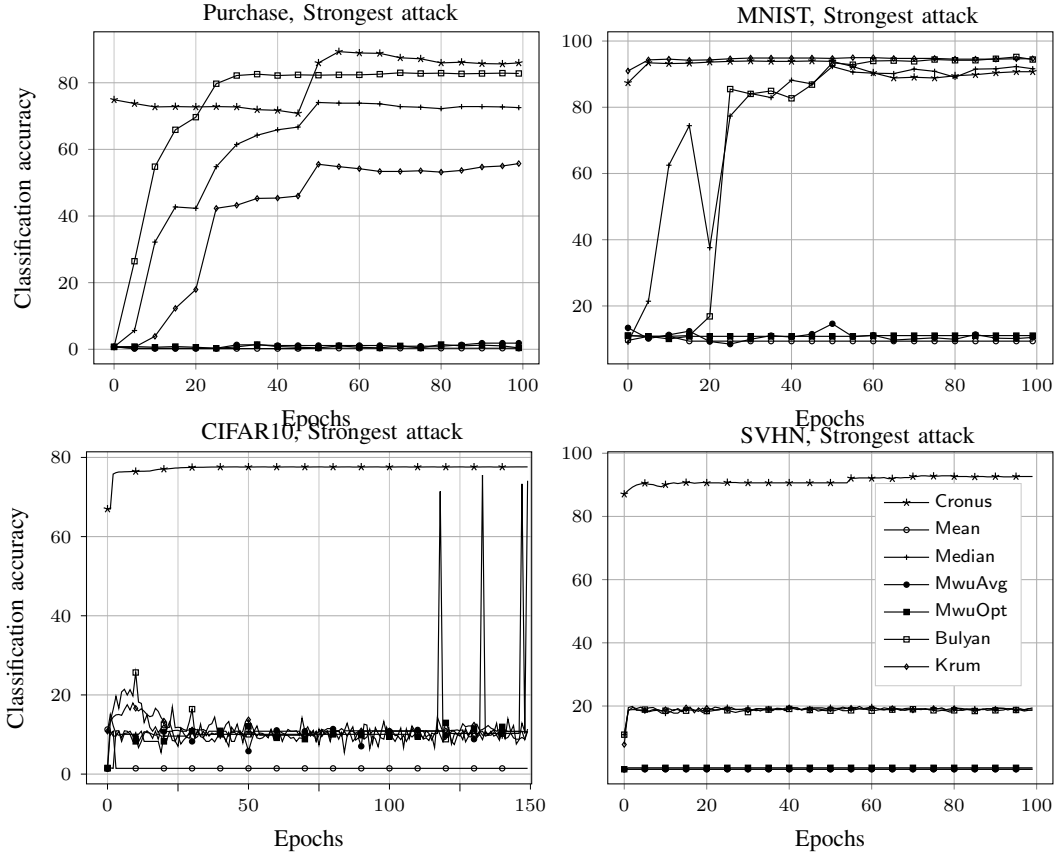

\vspace{-2em}
\centering
\begin{tabular}{cc}

\hspace{-3em}
\subfloat{\input{tex_figures/purchase_worse_attack_dab_1}}

&
\hspace{-2em}
\subfloat{\input{tex_figures/mnist_worse_attack}}

\\[-3ex]

\hspace{-3em}
\subfloat{\input{tex_figures/cifar10_worse_attack}}

&

\hspace{-2em}
\subfloat{\input{tex_figures/svhn_worse_attack}}
     
\end{tabular}
\vspace{-.7em}
\caption{
Convergence of Cronus and existing federated learning algorithms in \textbf{adversarial setting}.
Accuracy of Cronus in adversarial setting is almost the same as in benign setting (shown in Figure \ref{fig:benign_convergence}) due to its high robustness.
Except for CIFAR10, for which only collaboration phase is shown, both the Cronus training phases are shown in figure and the collaboration phase starts at epoch 50.
}

\label{fig:robustness_comparison}
\end{figure*}

Figure \ref{fig:robustness_comparison} shows the convergence of Cronus and federated learning.  We show the convergence plots of the algorithms in the benign setting, as a baseline for comparison, in Figure~\ref{fig:benign_convergence} in Appendix~\ref{appdx:exp_details}.  The federated learning algorithm works well in the absence of malicious parties, however, \textbf{all of the existing aggregation schemes in federated learning are significantly vulnerable to at least one poisoning attack.}

The weighted average aggregation (Section \ref{collaborative_learning}), i.e., FedAvg \cite{mcmahan2017communication}, is susceptible to all of the attacks:
\emph{The accuracy of FedAvg reduces close to random guess accuracy for all the datasets}.
$\mathsf{Median}$ aggregation is susceptible to PAF attack because the attack shifts the final aggregate along all the dimensions by a small amount to remain undetected, yet it can considerably damage the utility of the aggregated model.
Although $\mathsf{Bulyan}$ and $\mathsf{Krum}$ are robust aggregations, they are susceptible to the LIE attack.
As explained in Section \ref{attack:lie}, LIE attack exploits the sensitivity of  the parameters of neural networks to small perturbations.
The attack completely jeopardizes the accuracy of $\mathsf{Krum}$ aggregation, because the attack successfully forces $\mathsf{Krum}$ aggregation to select the malicious update as the aggregate in most of the epochs. Note that, the attack is not effective against MNIST classification task due to its simplicity, which allows the corresponding models to withstand the small perturbations.  $\mathsf{MwuAvg}$ and $\mathsf{MwuOpt}$ withstand all the attacks, but are susceptible to our OFOM attack, which we propose in Section~\ref{attack:our_attack}.
For all the datasets, the OFOM attack reduces the accuracy of the aggregations close to the random guess accuracy.

On the other hand, \textbf{the robustness of Cronus remains almost 1.0 as the classification accuracy of its models remains unaffected by any of the tested poisoning attacks}.
For the strongest attack on Cronus, the maximum reduction in accuracy is 0.4\% for Purchase, 1.3\% for SVHN, 1.5\% for MNIST, and 4.8\% for CIFAR10  models.
The reason for the high resilience of Cronus to the poisoning attacks is three-fold: 
First, as detailed in Section \ref{cronus:protocol}, reduced dimensionality of updates reduces the aggregate error and the sample complexities required to achieve the error.
Second, in Cronus, the models are not overwritten by aggregate models.  We instead take advantage of knowledge transfer algorithms~\cite{hinton2014distilling}.  The parameters of a model are sensitive to small perturbations which can prevent the model from converging \cite{baruch2019a}. 
Therefore, in the existing federated algorithms, the success of the poisoning attacks stems from the fact that a little noise introduced via the malicious parameter updates suffices to both prevent the model from convergence and to bypass the filtering of robust aggregation algorithms. Third, benign models have a stronger agreement on their predictions than on their parameter values. Thus, in Cronus, it is much easier to detect (or cancel the effect of) the poisoning vectors.





    

     


\subsection{Privacy}\label{exp:membership_risk}

We evaluate the risk of membership inference attacks on the participants' private training data during collaboration, and the effect of privacy preserving mechanisms \cite{abadi2016deep, nasr2018machine}.
As described in the threat model in Section \ref{ps:threat_model:meminf}, we assume the central server and other participants to run passive and active membership inference attacks \cite{nasr2019comprehensive} on individual party updates and their aggregates. 
We use Purchase, SVHN, and CIFAR10 datasets for our evaluation when 4 parties collaborate.

\subsubsection{Passive membership inference attacks} \label{exp:passive_mi_risk}

In the case of passive membership inference attacks, the server isolates the parties and mounts the attack separately on each of the collected updates, i.e., in case of FedAvg, attack is mounted on the parameter updates of each party and in case of Cronus, attack is mounted on the model obtained by training on the predictions shared by each party.
We also evaluate the privacy risk when the attack is mounted on the aggregate of these updates. 

The results are shown in Table \ref{tab:mi_risk}.
\textbf{The updates in FedAvg are highly susceptible to membership inference unlike the updates in Cronus.}
For the Purchase dataset, attack accuracy against the individual and aggregated updates in FedAvg is 78.1\% and 80.1\%, respectively,  whereas in Cronus they are 51.7\% and 51.9\%.
Unlike the prediction updates in Cronus, the high dimensional parameter updates in FedAvg encode a significantly higher amount of information about the party's local data.
Furthermore, knowledge transfer acts as a strong regularization method and mitigates the risk of membership inference attacks \cite{hinton2014distilling, song2018collaborative}.  It's important to note that knowledge transfer through predictions, on a dataset other than the training data, makes the behavior of the student model more indistinguishable on its training versus unseen data.  This happens as the distillation process does not carry the exceptionally distinguishable characteristics of the model on its training data, and results in smooth decision boundaries of the student model around the teacher's training data.  We observe similar results for SVHN and CIFAR10 datasets.

We also evaluate Cronus and FedAvg, when models use adversarial regularization during local training to improve membership privacy.
We note that, \textbf{adversarial regularization improves membership privacy of both FedAvg and Cronus}, but the increase is smaller for Cronus due to its inherent resilience to membership inference. For Purchase dataset with FedAvg, the  risk to individual and aggregated updates reduces by 9.9\% and 12.7\%, respectively, while for Purchase with Cronus, the risk to individual and aggregated updates is already very small, and it further reduces by 0.6\% and 1.1\%, respectively. 
Similarly for CIFAR10, privacy improvement in FedAvg is significantly more than in Cronus.
However, the privacy improvement for SVHN is very small even in case of FedAvg, due to large gaps in train and test accuracies at stronger adversarial regularization.

\subsubsection{Active membership inference attacks} \label{exp:active_mi_risk}

In each epoch, the server manipulates the aggregate update that it broadcasts to the parties by performing gradient ascent on the \emph{aggregated update} for a set of target data~\cite{nasr2019comprehensive}. 
In FedAvg, gradient ascent is performed directly on the aggregated parameters.  In Cronus, for running such attack, the server needs to train a model on the aggregated predictions while performing gradient ascent on the target data, and then, shares predictions of this model with the parties; we ensure that such model has accuracy close to the accuracy of party models in given epoch. 

\begin{figure*}
\vspace{-1em}
\centering
\begin{tabular}{ccc}
\hspace{-2em}
\subfloat{
\begin{tikzpicture}
\pgfmathsetlengthmacro\MajorTickLength{
      \pgfkeysvalueof{/pgfplots/major tick length} * 0.5
    }

\begin{axis}[
height=4.5cm,
width=6.7cm,
legend cell align={left},
legend style={draw=white!80.0!black,font=\scriptsize},
tick align=outside,
tick pos=left,
grid=both,
ticks=both,
title style={font=\small,at={(axis description cs:0.5,.95)}},
x grid style={lightgray!92.02614379084967!black},
xlabel={Epoch},
x label style={font=\small,at={(axis description cs:0.5,-0.13)}},
y label style={font=\small,at={(axis description cs:-0.07,.5)}},
xmin=-4.95, xmax=103.95,
xticklabel style={color=black,font=\scriptsize},
yticklabel style={color=black,font=\scriptsize},
minor ytick={2.5,7.5},
y grid style={lightgray!92.02614379084967!black},
ylabel={$\nabla\mathcal{L}(D_t)-\nabla\mathcal{L}(D)$},
ymin=-1, ymax=10,
major tick length=\MajorTickLength
]
\addplot [semithick, black, mark=*, mark size=1, mark repeat=5, mark options={solid}]
table {%
0 -0.0580000000000016
1 -0.0820000000000007
2 0.124
3 0.296
4 0.434
5 0.513
6 0.568
7 0.577
8 0.592
9 0.552
10 0.563
11 0.549
12 0.499
13 0.482
14 0.465
15 0.495
16 0.425
17 0.383
18 0.363
19 0.407
20 0.409
21 0.38
22 0.397
23 0.443
24 0.34
25 0.33
26 0.342
27 0.353
28 0.353
29 0.374
30 0.348
31 0.355
32 0.363
33 0.354
34 0.392
35 0.397
36 0.404
37 0.396
38 0.412
39 0.421
40 0.417
41 0.419
42 0.398
43 0.4
44 0.41
45 0.422
46 0.436
47 0.445
48 0.446
49 0.447
50 0.451
51 0.455
52 0.459
53 0.458
54 0.46
55 0.461
56 0.463
57 0.465
58 0.468
59 0.472
60 0.472
61 0.478
62 0.486
63 0.489
64 0.479
65 0.483
66 0.489
67 0.49
68 0.492
69 0.49
70 0.495
71 0.497
72 0.497
73 0.504
74 0.505
75 0.493
76 0.493
77 0.51
78 0.496
79 0.502
80 0.506
81 0.516
82 0.516
83 0.507
84 0.507
85 0.523
86 0.523
87 0.538
88 0.545
89 0.574
90 0.55
91 0.572
92 0.542
93 0.575
94 0.523
95 0.588
96 0.59
97 0.614
98 0.597
99 0.58
};
\addlegendentry{FedAvg}
\addplot [semithick, black, mark=square*, mark size=1, mark repeat=5, mark options={solid}]
table {%
0 0.1224
1 0.290813038581398
2 -0.00799999999999983
3 -0.0800090069761946
4 0.1966
5 0.187188173569659
6 0.0631999999999998
7 -0.0823854624213805
8 0.2722
9 0.176841501619179
10 -0.0759999999999998
11 -0.267321013613465
12 0.1418
13 0.0627902331394083
14 0.0896000000000001
15 0.158258226271447
16 0.1538
17 -0.00759895024688078
18 0.22
19 0.252756408782307
20 -0.00300000000000011
21 -0.0823908174759438
22 0.473800000000001
23 0.353854006967586
24 0.162
25 0.155473411863443
26 0.1448
27 0.145698826173976
28 0.0106000000000002
29 -0.0353208321400473
30 0.0527999999999999
31 0.038323696128618
32 -0.1364
33 -0.153421942610255
34 0.3932
35 0.345896618654945
36 0.3614
37 0.425573422096432
38 -0.0455999999999996
39 -0.200895879156541
40 -0.0592000000000006
41 -0.0943989435233718
42 0.0259999999999998
43 0.126402385453396
44 0.2918
45 0.299847783765235
46 0.2058
47 0.163443782237546
48 0.178599999999999
49 0.243271847464322
50 -0.0936
51 -0.173258475034299
52 0.218
53 0.282456168857135
54 0.1538
55 0.155372943814992
56 0.00219999999999985
57 -4.77865180165793e-05
58 0.2296
59 0.214391390770876
60 -0.1418
61 -0.0253253103165662
62 0.0135999999999996
63 0.0300466802902015
64 -0.3682
65 -0.314372728885299
66 0.0366
67 0.0677704367370218
68 0.1678
69 0.0688447830868277
70 0.0427999999999997
71 0.0875699286431178
72 -0.0436
73 -0.0930546628937115
74 -0.1074
75 -0.221179007350357
76 -0.0914000000000001
77 -0.0894324682253277
78 0.1412
79 0.10041296471544
80 0.0140000000000001
81 -0.0869185913596311
82 0.1006
83 0.204542442642671
84 0.0308
85 0.145724703289875
86 0.1004
87 0.0295939259805607
88 0.1432
89 0.042085824867133
90 0.0481999999999999
91 0.0576498367704545
92 0.0533999999999999
93 -0.0141077971860007
94 0.193
95 0.186020081061142
96 -0.0357999999999997
97 -0.0388584560266757
98 -0.0573999999999998
99 -0.126851540087444
};
\addlegendentry{Cronus}
\end{axis}

\end{tikzpicture} }

\hspace{-1em}
\subfloat{
\begin{tikzpicture}
\pgfmathsetlengthmacro\MajorTickLength{
      \pgfkeysvalueof{/pgfplots/major tick length} * 0.5
    }

\begin{axis}[
height=4.5cm,
width=6.7cm,
legend cell align={left},
legend style={draw=white!80.0!black,font=\scriptsize},
tick align=outside,
tick pos=left,
grid=both,
ticks=both,
title style={font=\small,at={(axis description cs:0.5,.95)}},
x grid style={lightgray!92.02614379084967!black},
xlabel={Epoch},
x label style={font=\small,at={(axis description cs:0.5,-0.13)}},
y label style={font=\small,at={(axis description cs:-0.07,.5)}},
xmin=-4.95, xmax=103.95,
xticklabel style={color=black,font=\scriptsize},
yticklabel style={color=black,font=\scriptsize},
minor ytick={2.5,7.5},
y grid style={lightgray!92.02614379084967!black},
ylabel={$\nabla\mathcal{L}(\bar{D}_t)-\nabla\mathcal{L}(\bar{D})$},
ymin=-1, ymax=10,
major tick length=\MajorTickLength
]
\addplot [semithick, black, mark=*, mark size=1, mark repeat=5, mark options={solid}]
table {%
0 -0.118
1 -0.129
2 -0.212
3 -0.138
4 -0.0960000000000001
5 -0.118
6 -0.0790000000000006
7 -0.0110000000000001
8 0.0330000000000004
9 0.149
10 0.247000000000001
11 0.349
12 0.684
13 0.63
14 0.707
15 0.721
16 0.701
17 0.831
18 0.811999999999999
19 0.619000000000001
20 0.941
21 1.048
22 1.222
23 1.086
24 1.304
25 1.54
26 1.688
27 1.792
28 1.876
29 1.881
30 1.861
31 2.034
32 2.129
33 2.111
34 2.046
35 2.254
36 2.368
37 2.349
38 2.293
39 2.488
40 2.545
41 2.565
42 2.528
43 2.684
44 2.753
45 2.764
46 2.658
47 2.827
48 2.922
49 2.937
50 2.945
51 2.899
52 2.804
53 2.709
54 2.741
55 2.697
56 2.67
57 2.678
58 2.663
59 2.662
60 2.707
61 2.702
62 2.718
63 2.767
64 2.772
65 2.809
66 2.816
67 2.873
68 2.919
69 2.921
70 2.99
71 2.995
72 3.063
73 3.113
74 3.182
75 3.286
76 3.372
77 3.417
78 3.515
79 3.589
80 3.65
81 3.786
82 3.774
83 3.869
84 3.918
85 4.004
86 4.055
87 4.153
88 4.131
89 4.139
90 4.208
91 4.302
92 4.45
93 4.396
94 4.559
95 4.547
96 4.521
97 4.632
98 4.621
99 4.655
};
\addplot [semithick, black, mark=square*, mark size=1, mark repeat=5, mark options={solid}]
table {%
0 -0.1272
1 -0.135931483020395
2 -0.2006
3 -0.174435540830435
4 0.0174
5 -0.0984878230130775
6 -0.0112000000000002
7 -0.0301878726987745
8 -0.2578
9 -0.383053690536617
10 -0.1588
11 -0.00776951304339875
12 -0.1038
13 -0.242696157752113
14 -0.1152
15 0.0262768503930651
16 -0.2088
17 -0.262879735526489
18 -0.1076
19 0.0391895419015675
20 -0.0100000000000001
21 0.0724247134414036
22 0.0810000000000002
23 0.148191903838888
24 -0.152999999999999
25 -0.282744796504061
26 -0.0520000000000003
27 -0.131517433726646
28 -0.1602
29 -0.147778118949513
30 -0.0856000000000002
31 -0.00296519958693438
32 -0.0921999999999997
33 -0.019728726717868
34 0.0389999999999993
35 -0.0250243533735713
36 0.1332
37 0.0367135707835033
38 -0.00640000000000001
39 -0.0693260409847937
40 0.1058
41 -0.00574305129740793
42 -0.120200000000001
43 -0.129646754873381
44 0.1474
45 0.136928539425112
46 0.0507999999999996
47 0.0945086347834739
48 -0.0374000000000002
49 -0.208266113207051
50 -0.1306
51 -0.180043075391718
52 -0.0694000000000003
53 -0.152182744631256
54 -0.1868
55 -0.275621984945171
56 -0.1446
57 -0.133181118163214
58 -0.1366
59 -0.081544520290943
60 -0.0492000000000004
61 0.000552388442158502
62 0.0291999999999994
63 -0.0296524321346801
64 -0.0463999999999999
65 0.0717556358777138
66 -0.1528
67 -0.0606264223897256
68 -0.0748000000000005
69 -0.121763321129077
70 -0.35
71 -0.333759117333144
72 0.0263999999999996
73 0.0471179027181996
74 0.171599999999999
75 0.200468299458995
76 0.0808
77 0.0129127242162433
78 -0.1204
79 -0.281727923209291
80 0.0138000000000005
81 -0.0241005633966857
82 -0.0850000000000001
83 -0.067756134167675
84 -0.000199999999999534
85 0.0526902475002373
86 -0.2556
87 -0.316117060019094
88 -0.1254
89 -0.137933433408904
90 -0.2224
91 -0.279040390953217
92 0.0124000000000002
93 0.0314366411820508
94 -0.1056
95 0.00243122737604082
96 0.0899999999999999
97 0.189080335957835
98 -0.1896
99 -0.203536211264878
};
\end{axis}

\end{tikzpicture} }

\hspace{-1em}
\subfloat{
\begin{tikzpicture}
\pgfmathsetlengthmacro\MajorTickLength{
      \pgfkeysvalueof{/pgfplots/major tick length} * 0.5
    }

\begin{axis}[
height=4.5cm,
width=6.7cm,
legend cell align={left},
legend style={at={(0.03,0.97)}, anchor=north west, draw=white!80.0!black,font=\scriptsize},
tick align=outside,
tick pos=left,
grid=both,
ticks=both,
title style={font=\small,at={(axis description cs:0.5,0.95)}},
x grid style={lightgray!92.02614379084967!black},
xlabel={Epoch},
x label style={font=\small,at={(axis description cs:0.5,-0.13)}},
y label style={font=\small,at={(axis description cs:-0.07,.5)}},
xmin=-4.95, xmax=103.95,
xticklabel style={color=black,font=\scriptsize},
yticklabel style={color=black,font=\scriptsize},
minor ytick={2.5,7.5},
y grid style={lightgray!92.02614379084967!black},
ylabel={$\nabla\mathcal{L}(\bar{D}_t)-\nabla\mathcal{L}(D_t)$},
ymin=-1, ymax=10,
major tick length=\MajorTickLength
]
\addplot [semithick, black, mark=*, mark size=1, mark repeat=5, mark options={solid}]
table {%
0 0.266
1 2.404
2 3.227
3 3.856
4 4.291
5 4.593
6 4.839
7 5.122
8 5.3
9 5.569
10 5.708
11 5.885
12 6.299
13 6.309
14 6.462
15 6.451
16 6.566
17 6.74
18 6.748
19 6.675
20 6.939
21 7.076
22 7.187
23 7.18
24 7.337
25 7.498
26 7.625
27 7.781
28 7.827
29 7.886
30 7.927
31 8.052
32 8.15
33 8.197
34 8.223
35 8.367
36 8.474
37 8.497
38 8.56
39 8.69
40 8.738
41 8.779
42 8.892
43 8.984
44 9.03
45 9.061
46 9.061
47 9.261
48 9.319
49 9.339
50 9.367
51 9.406
52 9.487
53 9.403
54 9.523
55 9.63
56 9.743
57 9.862
58 9.925
59 9.976
60 10.071
61 10.18
62 10.31
63 10.406
64 10.53
65 10.645
66 10.734
67 10.874
68 10.966
69 11.053
70 11.182
71 11.239
72 11.365
73 11.467
74 11.561
75 11.671
76 11.813
77 11.877
78 12.026
79 12.152
80 12.254
81 12.44
82 12.441
83 12.536
84 12.665
85 12.852
86 12.987
87 13.076
88 13.049
89 13.161
90 13.294
91 13.407
92 13.557
93 13.604
94 13.812
95 13.819
96 13.88
97 13.966
98 14.054
99 14.152
};
\addplot [semithick, black, mark=square*, mark size=1, mark repeat=5, mark options={solid}]
table {%
0 -0.1724
1 -0.326307559899296
2 -0.057
3 0.0338080976268959
4 -0.000400000000000134
5 -0.0320244965141402
6 0.0818000000000001
7 0.22014790906426
8 -0.2888
9 -0.261529590460401
10 0.0535999999999998
11 0.234915386617259
12 -0.0231999999999999
13 -0.128560545341498
14 -0.00620000000000012
15 0.0650182934498268
16 -0.1574
17 -0.0476264628229366
18 -0.1386
19 -0.0606849865784021
20 0.17
21 0.283432247347316
22 -0.25
23 -0.140691140370122
24 -0.142599999999999
25 -0.0958951669366122
26 -0.0294000000000004
27 -0.044851014406742
28 -0.0293999999999997
29 0.0320276667829617
30 0.0269999999999996
31 0.11892174923338
32 0.1664
33 0.125706061520704
34 -0.162200000000001
35 -0.166008016464537
36 0.0224000000000004
37 0.00615059362136421
38 0.0991999999999997
39 0.172171651387853
40 0.3876
41 0.37954276328264
42 0.0259999999999998
43 0.01650163288611
44 0.0536000000000001
45 0.0449208376344643
46 -0.0596000000000004
47 0.00402738829860496
48 -0.0573999999999998
49 -0.188316904375845
50 0.0260000000000005
51 0.0309561352494008
52 -0.1788
53 -0.211326069610439
54 -0.1818
55 -0.252537314318051
56 -0.0201999999999998
57 -0.0153455024496836
58 -0.2366
59 -0.196236262344394
60 0.2236
61 0.204704143083004
62 0.1226
63 0.103606470999406
64 0.4112
65 0.481071296010064
66 -0.0332000000000001
67 -0.0345790522812521
68 -0.1562
69 -0.111171437022934
70 -0.206
71 -0.169936596536952
72 0.2018
73 0.195312652797436
74 0.390799999999999
75 0.384351319619088
76 0.252800000000001
77 0.237323207291882
78 -0.212
79 -0.295138808949442
80 0.0448
81 0.0518341387486728
82 -0.1032
83 -0.0541158881072079
84 0.121400000000001
85 0.0808526080101473
86 -0.2962
87 -0.363609059911813
88 -0.2114
89 -0.141545891467167
90 -0.2
91 -0.304440209358805
92 0.0859999999999999
93 0.129036379460372
94 -0.1896
95 -0.153463270638628
96 0.222
97 0.178156737390596
98 -0.0765999999999998
99 0.070299171007121
};
\end{axis}

\end{tikzpicture} }
     
\end{tabular}
\caption{Difference in the gradient-norms, $\nabla\mathcal{L}(D)$, of the last layer of aggregated model on the target and non-target data (Purchase100 data). In the context of active membership inference attacks, 
$D$, $D_t$, $\bar{D}$, and $\bar{D}_t$ denote non-target members, target members, non-target non-members, and target non-members, respectively.
}

\label{fig:grad_ascent}
\end{figure*}
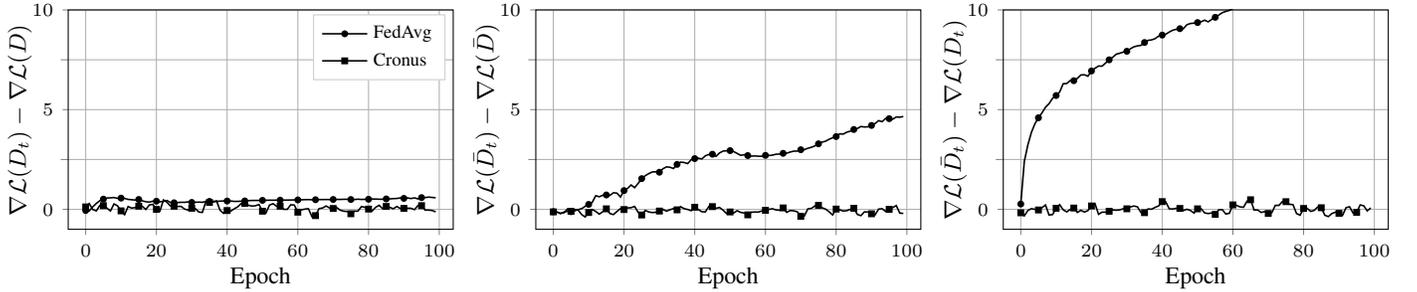

Table \ref{tab:mi_risk} shows the results.  \textbf{The active attacks significantly increase the privacy risk of the target data in case of FedAvg}: for Purchase dataset, the risk due to individual update increases by 7.8\% (77.1\% to 84.9\%), while due to aggregated update increases by 8\% (74.7\% to 82.7\%).
But, \textbf{in case of Cronus, the active attacks are ineffective and the increase in risk is negligible}: 0.3\% for individual update while 1.1\% for aggregated update.
In Figure \ref{fig:grad_ascent}, we show the effect of gradient ascent on the difference in gradient-norms of target and non-target data for aggregated model for the Purchase dataset. This directly correlates with the success of membership inference \cite{nasr2019comprehensive}.  We observe that, for FedAvg on SVHN dataset, the active attacks increase the risk to individual and aggregate updates by 2.5\% and 4.4\%, respectively, but, the increased risk negligible for Cronus.

\begin{table*}
\fontsize{8}{11}\selectfont{}
\caption{
Accuracy of passive and active membership inference attacks with central server as adversary.
We also evaluate effect of adversarial regularization (with parameter $\lambda$) used to preserve membership privacy.
We use 4 parties and data per party as in Table \ref{tab:exp_setup}.
}

\label{tab:mi_risk}
\begin{center}
\vspace{-1em}
\begin{tabular}{| c |  c | c | c | c | c | }
\hline

\multirow{3}{*}{Dataset} & Federated & \multicolumn{2}{c|}{Attack on party update} & \multicolumn{2}{c|}{Attack on aggregated update}\\ \cline{3-6}
& learning & Passive & Active &  Passive & Active \\
& algorithm & attack acc. & attack acc. & attack acc. & attack acc. \\ 
\hhline{|=|=|=|=|=|=|}

Purchase & FedAvg & 77.1 & 84.9 & 74.7 & 82.7 \\ \cline{3-6}
(without privacy) & Cronus & 54.6 & 54.9 & 53.6 & 54.7 \\ 
\hhline{|=|=|=|=|=|=|}

Purchase & FedAvg & 70.8 & 77.3 & 69.9 & 77.0 \\ \cline{3-6}
(with membership privacy, $\lambda=3$) & Cronus & 54.1 & 51.5 & 53.7 & 54.6 \\ 
\hhline{|=|=|=|=|=|=|}

SVHN & FedAvg & 64.8 & 67.3 & 59.9 & 64.3 \\ \cline{3-6}
(without privacy) & Cronus & 55.6 & 53.1 & 56.5 & 55.7 \\ 
\hhline{|=|=|=|=|=|=|}

SVHN & FedAvg & 64.9 & 67.0 & 60.0 & 64.2 \\ \cline{3-6}
(with membership privacy, $\lambda=0.5$) & Cronus & 54.1 & 56.9  & 55.6 & 55.0 \\ 
\hhline{|=|=|=|=|=|=|}

CIFAR10 & FedAvg & 79.9 & 80.5 & 76.8 & 77.1\\ \cline{3-6}
(without privacy) & Cronus & 57.0 & 57.8 & 55.5 & 56.7\\
\hhline{|=|=|=|=|=|=|}

CVIFAR10 & FedAvg  & 59.9 & 64.1 & 59.6 & 62.2 \\ \cline{3-6}
(with membership privacy, $\lambda=1$) & Cronus & 52.9 & 54.4 & 52.6 & 57.0\\ 
\hline

\end{tabular}
\end{center}

\end{table*}

\subsection{Differential privacy}\label{exp:fed_dp}

We compare the performance of Cronus and the conventional federated learning with user-level~\cite{mcmahan2018learning} or record-level~\cite{abadi2016deep} differential privacy.
For both of these comparisons, we use SVHN dataset with 32 benign parties and the model architecture as described in Table \ref{tab:exp_setup}, and use the moments accountant method (and the code)~\cite{abadi2016deep, tf_privacy} to bound the total privacy risk. Note that we train the {\em whole model} using differential privacy, as opposed to only training the last layer~\cite{abadi2016deep}.  Our results shows the significant accuracy cost of some existing DP algorithms for federated learning, and calls for designing DP algorithms with high utility and tight privacy loss.

\subsubsection {User-level DP \cite{mcmahan2018learning}}

The user-level DP (UDP-FedAvg) algorithm proposed by McMahan et al. \cite{mcmahan2018learning} cannot lead to training good accuracy models, when the number of parties in each epoch is small.  Even for large privacy budgets, i.e., $\epsilon = 100$, the parties do not benefit from collaboration, and with the user-level DP, the global model in FedAvg achieves close to random-guess accuracy.  We observed similar results for running user-level DP on Cronus with small number of participants. This result is expected as the sensitivity of the element-wise mean aggregation algorithm is inversely proportional to the number of parties, and a very large number of parties is required to reduce the noise, e.g., \cite{mcmahan2018learning} uses $5000$ parties in each epoch.

\subsubsection{Record-level DP \cite{abadi2016deep}}

We empirically show that the conventional parameter aggregation in FedAvg is not suitable to provide the record-level DP, and is also susceptible to poisoning attacks.

The results are shown in Table \ref{tab:record_level_dp} for SVHN dataset.  The accuracy of the models for federated learning or Cronus, with DP-SGD, cannot reach the accuracy of stand-alone training, which makes the collaboration useless. The results of the strongest poisoning attacks show that DP-SGD FedAvg has no robustness against the attacks.

\begin{table}
\fontsize{8}{11}\selectfont{}
\caption{
Accuracy and robustness of models for record level DP \cite{abadi2016deep} with $\epsilon=15.4$ on the SVHN dataset.  The baseline stand-alone accuracy is 87\%. 
}

\label{tab:record_level_dp}
\vspace{-1em}
\begin{center}
\setlength{\extrarowheight}{-0.03cm}

\begin{tabular}{| c || c | c || c | c || c | c |}
\hline
{\# of } & \multicolumn{2}{c||}{Accuracy (Benign)} & \multicolumn{2}{c||}{Worst accuracy} & \multicolumn{2}{c|}{Robustness}\\ \cline{2-7}
parties & FedAvg & Cronus & FedAvg & Cronus & FedAvg & Cronus \\  \hline
32 & 65.7 & 85.8 & 4.5 & 83.1 & 0.07 & \textbf{0.97} \\
16 & 74.3 & 84.8 & 8.1 & 84.0& 0.11 & \textbf{0.99}\\
8 & 77.9 & 84.2 & 0.9 & 82.2 & 0.01 & \textbf{0.98}\\
4 & 81.6 & 81.5 & 1.7 & 81.2 & 0.02 & \textbf{0.98} \\ \hline

\end{tabular}
\end{center}

\end{table}

\subsection{Cronus with heterogeneous model architectures}\label{exp:hetero}

Due to the use of predictions based updates, Cronus allows parties with heterogeneous model architectures to participate in collaboration.
Below, we compare different aspects of  the homogeneous and heterogeneous collaborations.  We use Purchase data and 5 fully connected models, which we call  A1, A2, A3, A4, and A5, with hidden layer sizes {\{\}}, {\{1024\}}, {\{512, 256\}}, {\{1024, 256\}}, and {\{1024, 512, 256\}} respectively.
Note that, A1 models, called \emph{bad models}, have lower capacity and accuracy than A2-5 models, which we call \emph{good models}.
We denote by  $P_{j:k}$ the model architectures of parties $\in [j,k]$.
We denote the entire collaboration in curly brackets, e.g., we denote the collaboration of 3 sets of 4 models, i.e. 12 models in total, each with either of A3, A3, or A4 models by \{$P_{1:4}$ = A2, $P_{5:8}$ = A3, $P_{9:12}$ = A4\}.
In tables, accuracy of an architecture is the average accuracy of all the models with that architecture, e.g., in Table  \ref{tab:hetero_hetero} accuracy of A2 is average of accuracies of all models with A2 architecture in the two collaborations.

First, we show that the heterogeneous collaboration between models of equivalent capacities does not reduce the accuracy of party models compared to its homogeneous counterparts.
We consider four homogeneous collaborations each of 16 parties such that \{$P_{1:16}$ = A2\}, \{$P_{1:16}$ = A3\}, \{$P_{1:16}$ = A4\}, and \{$P_{1:16}$ = A5\}, and compare it with a heterogeneous collaboration: \{$P_{1:4}$ = A2, $P_{5:8}$ = A3, $P_{9:12}$ = A4, $P_{13:16}$ = A5\}.
The results are shown in Table \ref{tab:homo_hetero}.

Next, we show that \textbf{the presence of a few bad models does not affect the accuracy of the good models in the heterogeneous collaboration, while significantly benefits the bad models}.
Specifically, we show that accuracy of the collaboration of 12 good models, i.e., \{$P_{1:4}$ = A2, $P_{5:8}$ = A3, $P_{9:12}$ = A4\} remains unaffected even if 4 bad models are added to it, i.e., $P_{13:16}$ = A2, as shown in Table \ref{tab:hetero_hetero}.

Finally, we show that \textbf{heterogeneity allows for more knowledge sharing via collaboration and always improves the utility of collaborations}.
We consider 4 homogeneous collaborations: \{$P_{1:4}$ = A1\}, \{$P_{1:4}$ = A2\}, \{$P_{1:4}$ = A3\}, and \{$P_{1:4}$ = A4\} and compare them with a heterogeneous collaboration that includes all these 16 parties, i.e., \{$P_{1:4}$ = A1, $P_{5:8}$ = A2, $P_{9:12}$ = A3, $P_{13:16}$ = A4\}.
Table \ref{tab:more_hetero}  shows that including more participants clearly benefits all types of models, although the bad models benefit more than the good ones. 
For instance, A1 models improve by 8\% from 70.1\% to 78.1\% due to heterogeneous collaboration, while A2, A3, and A4 models improve by 3.2\%, 2.2\%, and 1.4\%, respectively.

\begin{table}[h]
\begin{center}
\setlength{\extrarowheight}{-0.03cm}
\fontsize{8}{11}\selectfont{}
\caption{ Comparison between heterogeneous and homogeneous collaborations in Cronus.}
\label{tab:homo_hetero}
\hspace{-.5em}
\begin{tabular}{| c | c | c | c || c |}
\hline
\multicolumn{4}{|c||}{Homogeneous} & Heterogeneous\\ \hline
\multirow{2}{*}{$P_{1:16}\rightarrow$ A2} & \multirow{2}{*}{A3} & \multirow{2}{*}{A4} & \multirow{2}{*}{A5} & \{$P_{1:4}$ = A2, $P_{5:8}$ = A3\\
  &  &  &  & $P_{9:12}$ = A4, $P_{13:16}$ = A5\} \\  \hline

89.6 & 89.3 & 88.4 & 88.6 & 89.3\\ \hline
\end{tabular}

\vspace{1em}

\caption{ Effect of the presence of low accuracy bad models on the performance of higher accuracy good models. $n$ is the number of collaborating parties.}
\label{tab:hetero_hetero}

\begin{tabular}{| c | c | c  |}

\hline

\multirow{3}{*}{Models} & Heterogeneous & Heterogeneous \\ \cline{2-3}
& \{$P_{1:4}$ = A2, $P_{5:8}$ = A3, & \{$P_{1:4}$ = A2, $P_{5:8}$ = A3, \\
& $P_{9:12}$ = A4\} & $P_{9:12}$ = A4, $P_{13:16}$ = A1\} \\ \hline
A1 &  - & 78.1 \\
A2 & 88.5 & 88.7 \\
A3 & 88.6 & 88.1 \\
A4 & 88.7 & 88.1 \\

\hline
\end{tabular}

\vspace{1em}

\caption{More participation due to heterogeneity always improves the overall utility of the collaboration.}
\label{tab:more_hetero}

\begin{tabular}{| c | c | c  |}

\hline

\multirow{3}{*}{Models} & Homogeneous & Heterogeneous \\ \cline{2-3}
& 4 small collaborations & \{$P_{1:4}$ = A2, $P_{5:8}$ = A3, \\
& {$P_{1:4}$ = A1/A2/A3/A4} & $P_{9:12}$ = A4, $P_{13:16}$ = A1\} \\ \hline

A1 &  70.1 & 78.1 \\
A2 & 85.5 & 88.7 \\
A3 & 85.9 & 88.1 \\
A4 & 86.7 & 88.1 \\ \hline

\end{tabular}

\end{center}
\end{table}

\begin{figure}[t]
 	\centering
 	
 	\input{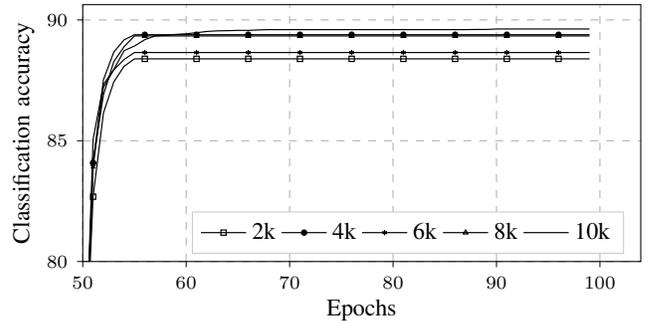}
 	\vspace{-.5em}
 	\caption{Communication reduction by randomly subsampling from the 10k public data in each epoch of Cronus collaboration phase.
 		The final Cronus accuracy is not affected even when just 2k samples are used in each Cronus epoch.
 	}
 	
 	\label{fig:compression_subsample}
 	\vspace{-1em}
\end{figure}

\subsection{Communication overhead of Cronus}

Cronus significantly reduces the communication due to sharing of  predictions instead of parameters of party models.
 Now, we demonstrate that using random subset of public data in each epoch further reduces communication without compromising the accuracy of Cronus.
 In Figure \ref{fig:compression_subsample}, we show the \emph{Cronus training progress in its collaboration phase} when public data of sizes 2k, 4k, 6k, 8k, 10k is sampled randomly in each epoch; the corresponding accuracies are 88.4, 89.4, 88.7, 89.3, 89.6, respectively.
 Knowledge transfer improves convergence of student models \cite{anil2018large}, and therefore, sub-sampling does not reduce the convergence rate of Cronus.
 Therefore, Cronus does not only reduce the per-epoch communication, but also reduces the overall communication.

\section{Related work}

Federated learning aims at training machine learning models via collaboration, without sharing data among the participants~\cite{shokri2015privacy}.  However, existing federated learning algorithms are susceptible to various poisoning attacks which deters data holders from trusting the algorithm~\cite{bhagoji2019analyzing, bagdasaryan2018how, melis2019exploiting}.  Bagdasaryan et al.  \cite{bagdasaryan2018how} demonstrate a targeted poisoning attack against FedAvg \cite{mcmahan2017communication} with just one malicious party.
Bhagoji et al. \cite{bhagoji2019analyzing} also propose a targeted attack with a single malicious party to work against two robust aggregation schemes, Krum \cite{blanchard2017machine} and coordinate-wise median \cite{yin2018byzantine}. 

To address the poisoning attacks, many robust aggregation schemes are proposed in the literature (\cite{mhamdi2018the,blanchard2017machine,yin2018byzantine}). 
Yin et al.~\cite{yin2018byzantine} analyze median and trimmed-mean aggregation rules with provable robustness guarantees.  Median and geometric-median based robust aggregation rules are also extensively explored \cite{xie2018generalized, su2017defending, alistarh2018byzantine}. Blanchard et al. \cite{blanchard2017machine} propose Krum aggregation, a computationally tractable modification of geometric-median, which selects the update that is closest to its neighbors as the aggregate in a given epoch. El Mhamdi et al.~\cite{mhamdi2018the} propose Bulyan, a two-step meta-aggregation algorithm based on the Krum and trimmed median, which filters malicious updates followed by computing the trimmed median of the remaining updates. In spite of their robustness guarantees, these algorithms are shown to be susceptible to targeted or generic poisoning attacks against deep learning models~\cite{baruch2019a, bhagoji2019analyzing}.  

There is a long line of research that considers the problem of robust mean estimation of a corrupted set of data samples \cite{diakonikolas2016robust,diakonikolas2017being,diakonikolas2019sever,li2018principled}, and proposes algorithm to tighten the error bounds and reduce its dependence on the data dimensionality.
The error rate of most of these aggregation algorithms depends on the dimensionality of their inputs \cite{lai2016agnostic,diakonikolas2017being,diakonikolas2016robust}.
Therefore, the high dimensional updates in federated deep learning setting significantly reduces the effectiveness of these algorithms (as their error bounds in presence of poisoning large vectors are lose). 
In summary, the high dimensional updates hinder the use of sophisticated robust algorithms and also enable strong poisoning attacks.  Our Cronus algorithm, inspired by the knowledge transfer algorithms \cite{ba2014deep, hinton2014distilling}, addresses these concerns by sharing the knowledge of the local party models via their predictions on a non-sensitive public data. This novel knowledge sharing federated learning scheme significantly reduces the dimensions of updates, and therefore, reduces the aggregation error.

Although federated learning prevents its participants from sharing  their private data, recent works~\cite{nasr2019comprehensive, orekondy2018understanding, melis2019exploiting} demonstrate passive and active inference attacks against this setting, and successfully infer sensitive information about the parties' private data.  Such inference  attacks can be defended by using differentially private (DP) learning \cite{abadi2016deep, mcmahan2018learning, dwork2018privacy}.  By trusting the aggregator, McMahan et al.~\cite{mcmahan2018learning} consider a federated setting where the server collects clipped gradient updates from the parties and then adds a calibrated noise using bounded sensitivity estimators; however, this algorithm can achieve acceptable accuracy only with a large number of participants. Besides, it does not protect data privacy against the aggregator. 
Treux et al.~\cite{treux2018a} propose a hybrid approach for privacy-preserving federated learning that leverages DP and secure multi-party computation among collaborators.  These works, although provably privacy-preserving, are not robust to poisoning attacks and produce  models with undesirably poor privacy-utility tradeoffs \cite{jayaraman2019evaluating}.

Knowledge of ensemble of teacher models has been used to train a student model in a few previous works~\cite{hamm2016learning, papernot2017semi, papernot2018scalable, anil2018large, jeong2018communication}.  Papernot et al. \cite{papernot2017semi} propose PATE, a centralized learning approach that uses ensemble of teachers to label a subset of unlabeled non-sensitive public data, and then, trains a student in a semi-supervised fashion with differential privacy~\cite{salimans2016improved}.  PATE's setting is fundamentally different from that of Cronus in that all the teacher models and the noise generation mechanisms in PATE are performed by a trusted entity who owns all the data.  Co-distillation approach is a method to use distillation on private training data with other parties~\cite{anil2018large, jeong2018communication}, however, without defending against data poisoning and inference attacks~\cite{shokri2017membership, yeom2018privacy, hayes2019logan}.
\section{Conclusions}

We have proposed Cronus learning algorithm as a variant of federated learning to mitigate the three fundamental limitations of federated learning: models are susceptible to data/parameter poisoning attacks, and inference attacks against local training data, and cannot operate on models with heterogeneous architectures.  We have used the predictions of the local models, instead of their parameters, to exchange the knowledge between the local private models.  This enables heterogeneous federated learning. It also significantly reduces the dimensionality of inputs of aggregation algorithms, which results in tight robustness guarantees even for networks with a small number of participants. Distilling the models' knowledge, through their prediction vectors, completely eliminates the possibility of white-box inference attacks, reduces the overall leakage about local datasets, and enables using utility-preserving mechanisms for prediction privacy.

{
\bibliographystyle{IEEEtranS_7}
\bibliography{privacy}
}

\newpage
\appendix
\section{Appendix}

\begin{table*}

\fontsize{8}{11}\selectfont{}
\begin{center}

\caption{
Evaluation of the conventional federated learning with various aggregation schemes with Cronus learning using the strong poisoning attacks described in Section \ref{attacks}.
Robustness in Table \ref{tab:fedlearn} is measured as the ratio of the accuracy of the final model/s when the strongest attack is mounted  and the accuracy in the benign setting; the strongest attack is determined empirically as the one that maximally reduces the accuracy of the corresponding federated learning aggregation.
}

\label{tab:fedlearn_all}

\begin{tabular}{ c | c |c |c |c |c |c |c |c |}
\cline{2-9}

\multirow{2}{*}{Dataset} &  & \multicolumn{6}{c|}{Federated learning with various aggregation algorithms} & \multirow{2}{*}{$\mathsf{Cronus}$} \\ \cline{3-8}

&  & $\mathsf{Mean}$ & $\mathsf{Median}$ & $\mathsf{MwuAvg}$ & $\mathsf{MwuOpt}$ & $\mathsf{Bulyan}$ & $\mathsf{Krum}$ &  \\ \hline
\hline

\multirow{5}{*}{SVHN} & Accuracy (Benign) &  \cellcolor[gray]{0.9}\textbf{95.9} & \cellcolor[gray]{0.9}94.8 & \cellcolor[gray]{0.9}93.9 & \cellcolor[gray]{0.9}94.4 &  \cellcolor[gray]{0.9}94.5 & \cellcolor[gray]{0.9}89.6 & \cellcolor[gray]{0.9}91.1\\ 
 & Label flip & 92.9 & 90.1 & 91.2 & 89.3 &  93.9 & 88.6 & \cellcolor[gray]{0.8}89.8\\ 
 & LIE & 14.8 & \cellcolor[gray]{0.8}14.5 & 91.6 & 92.0 &  \cellcolor[gray]{0.8}15.5 & \cellcolor[gray]{0.8}16.2 & 91.5 \\ 
 & OFOM & \cellcolor[gray]{0.8}0.9 & 94.5 & \cellcolor[gray]{0.8}0.9 & \cellcolor[gray]{0.8}0.7 & 94.4 &  89.0 & 91.0\\ 
 & PAF & 12.8 & 16.4 & 95.1 & 93.1 &  93.4 & 87.5 &  91.1 \\ \hline\hline

\multirow{5}{*}{MNIST} & Accuracy (Benign) & \cellcolor[gray]{0.9}96.7  & \cellcolor[gray]{0.9}96.5 & \cellcolor[gray]{0.9}\textbf{97.2} & \cellcolor[gray]{0.9}97.4 & \cellcolor[gray]{0.9}96.9 & \cellcolor[gray]{0.9}93.3 & \cellcolor[gray]{0.9}95.2\\ 
 & Label flip & 96.3 & 94.4 & 94.7 & 93.6 & 96.8 & \cellcolor[gray]{0.8}89.9 & \cellcolor[gray]{0.8}95.0\\ 
 & LIE & 95.1  & 93.1  & 95.6  & 96.7  & \cellcolor[gray]{0.8}94.1 & 94.3 & 95.9 \\
 & OFOM & 22.1 & 97.3 & \cellcolor[gray]{0.8}25.3 & 36.0 & 97.1 & 94.4 & 96.1 \\
 & PAF & \cellcolor[gray]{0.8}9.6 & \cellcolor[gray]{0.8}91.5 & 96.9 & \cellcolor[gray]{0.8}12.7 & 97.1 & 94.0 & 96.2 \\
\hline
\hline

\multirow{5}{*}{Purchase} & Accuracy (Benign) & \cellcolor[gray]{0.9}93.3  & \cellcolor[gray]{0.9}93.0 & \cellcolor[gray]{0.9}\textbf{93.6} & \cellcolor[gray]{0.9}92.5 & \cellcolor[gray]{0.9}92.8 & \cellcolor[gray]{0.9}72.1 & \cellcolor[gray]{0.9}89.6 \\ 
 & Label flip & 88.9 & 89.9 & 63.4 & 67.6 & 91.7 & 74.8 & \cellcolor[gray]{0.8}88.0 \\ 
 & LIE & 2.5 & 69.3 & 92.2 & 85.6 & \cellcolor[gray]{0.8}81.8 & \cellcolor[gray]{0.8}49.6 & 89.2 \\
 & OFOM & 1.4 & 92.8 & \cellcolor[gray]{0.8}1.8 & \cellcolor[gray]{0.8}1.1 & 92.6 & 74.5 & 89.4 \\
 & PAF & \cellcolor[gray]{0.8}1.1 & \cellcolor[gray]{0.8}12.5 & 93.0 & 88.0 & 91.0 & 76.6 & 89.4\\
\hline
\hline

\multirow{4}{*}{CIFAR10} & Accuracy (Benign) & \cellcolor[gray]{0.9}88.4 & \cellcolor[gray]{0.9}\textbf{89.1} & \cellcolor[gray]{0.9}86.2 & \cellcolor[gray]{0.9}87.6 &  \cellcolor[gray]{0.9}89.0 & \cellcolor[gray]{0.9}84.5 & \cellcolor[gray]{0.9}80.1 \\ 

 & Label flip & -- & -- & -- & -- &  -- & -- & 79.8\\ 
 & LIE & 18.9 & 61.2 & 86.0 & 84.3 &  \cellcolor[gray]{0.8}75.6 & \cellcolor[gray]{0.8}18.0 & \cellcolor[gray]{0.8}78.0 \\ 
 & OFOM & 12.9 & 89.5 & \cellcolor[gray]{0.8}14.2 & \cellcolor[gray]{0.8}12.8 & 89.1 &  85.0 & 78.5\\ 
 & PAF & \cellcolor[gray]{0.8}11.3 & \cellcolor[gray]{0.8}15.1 & 86.4 & 85.0 &  89.0 & 839 & 79.0 \\ 

 \hline

\end{tabular}
\end{center}

\end{table*}

\subsection{Details datasets and model architectures}\label{appdx:setup}

We use four datasets in our evaluation, whose details follow.

\paragraphb{SVHN.} 
SVHN \cite{netzer2011reading} dataset contains Google's street view images of house numbers.
The images are 32x32, with 3 floating point numbers containing the RGB color information of each pixel.
We use the extended SVHN dataset with 630,420 samples to train 32 party models each with 5,000 training samples; the public data size is 10,000.
We use validation and test data of sizes 2,500 each.
The reference data required for adversarial regularization is of the same size as that of training data for the cases of all the datasets.

\paragraphb{MNIST.} 
MNIST \cite{lecun1998gradient} dataset contains 28x28 images of handwritten digits and is composed of 60,000 training samples and 10,000 test samples.
The dataset contains 10 classes each with 60,000 training and 1,000 test samples.
We use validation and test data of sizes 1,000 each.
We use 28 parties each with 2,000 training and reference samples, and public data size is 10,000.

\paragraphb{Purchase.} 
Purchase \cite{purchase} dataset contains the shopping records of several thousand online customers, extracted during Kaggle's Acquire Valued Shopper challenge \cite{purchase}. 
The dataset contains 197,324 data records with feature vectors of 600 dimensions and corresponding class label from one of total 100 classes.
We use validation and test data of sizes 2,500 each.
We use 16 parties each with 10,000 training and reference data, and public data size is 10,000.

\paragraphb{CIFAR10.}
CIFAR10 \cite{krizhevsky2009learning} has 60,000 color (RGB) images (50,000 for training and 10,000 for testing), each of 32 $\times$ 32 pixels.
The images are clustered into 10 classes based on the objects in the images and each class has 5,000 training and 1,000 test images.
We use validation and test data of sizes 2,500 each.
We use 16 parties each with 2,500 training data, and public data size is 10,000.

\paragraphb{Model architectures.}
For SVHN, we use a neural network with three convolution layers and one fully connected layer, and Relu activations.
For the MNIST dataset, we use a fully connected neural network with layer sizes \{784, 256, 64, 10\} and Relu activations.
For the Purchase dataset, we use fully connected neural networks with layer sizes \{600, 1024, 100\} and Tanh activations.
For CIFAR10 dataset, we use DenseNet architecture \cite{huang2017densely} with 100 layers and growth rate of 12.
For the heterogeneity experiments with Purchase dataset, we use 5 fully connected networks with hidden layer sizes [\{\},\{1024\},\{512, 256\},\{1024, 256\}, \{1024, 512, 256\}]; here \{\} implies that the corresponding model has no hidden layers.

\paragraphb{Training hyper-parameters.}
The initialization and collaboration phases of SVHN, MNIST, and Purchase trained models are of 50 epochs each.
In both the phases, we train party models on their local training data  using Adam optimizer at 0.0005 learning rate.
Additionally, in collaboration phase, i.e., for epochs 50-100, we train party models on public data, $(X_p,\bar{Y})$, using SGD optimizer at  a learning rate of 0.001.
For CIFAR10, we train models for 200 epochs using SGD optimizer with 0.1 learning rate, 0.9 momentum, and $10^-4$ weight decay in both the phases.
Additionally, in collaboration phase, we train the models on public data using SGD optimizer at 0.01 learning rate, 0.99 momentum, and $10^{-6}$ weight decay.

For experiments of membership privacy assessment, we use state-of-the-art whitebox inference model proposed in \cite{nasr2019comprehensive}, and  use the gradients and outputs of its last layer, in addition to the blackbox access features including prediction of input and its cross-entropy loss.
We train the inference model using Adam optimizer at a learning rate of 0.0001 for 100 epochs.








\begin{figure*}
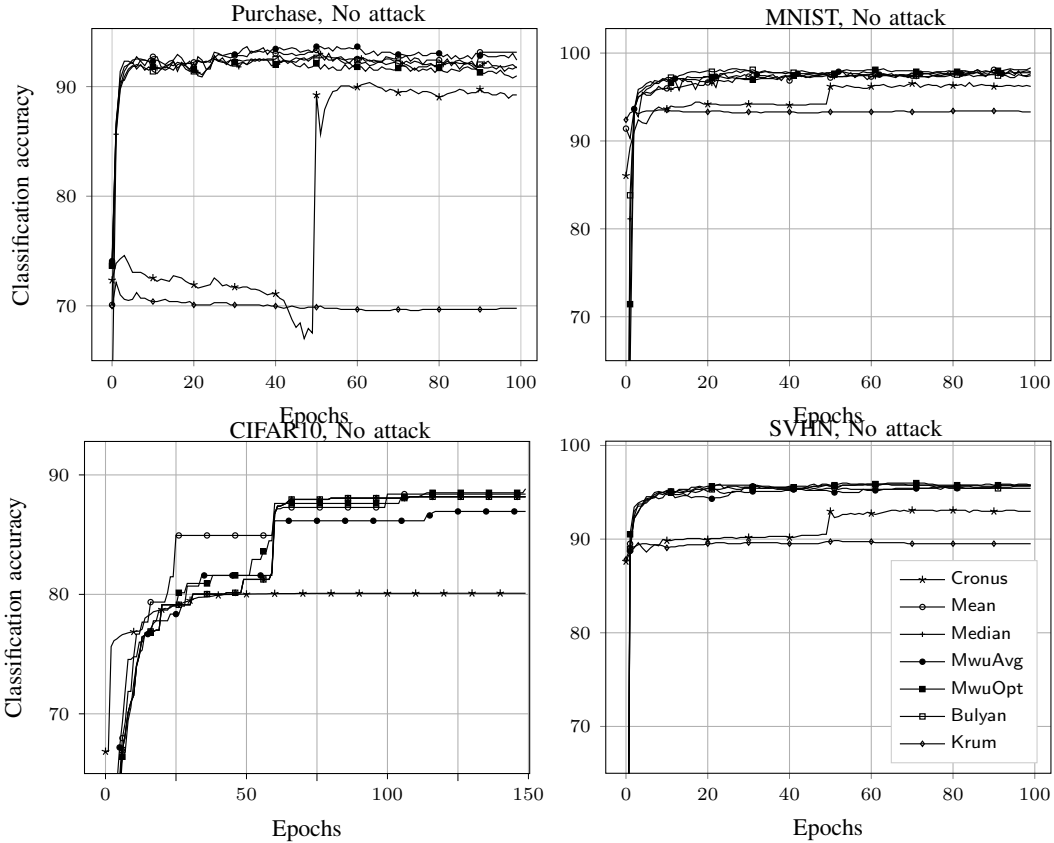

\vspace{-2em}
\centering
\begin{tabular}{cc}

\hspace{-3em}
\subfloat{\input{tex_figures/purchase_no_attack_dab_1}}

&
\hspace{-2em}
\subfloat{\input{tex_figures/mnist_no_attack}}

\\[-3ex]

\hspace{-3em}
\subfloat{\input{tex_figures/cifar10_no_attack}}

&

\hspace{-2em}
\subfloat{\input{tex_figures/svhn_no_attack}}
     
\end{tabular}
\vspace{-.7em}
\caption{
Convergence of Cronus and existing federated learning algorithms in \textbf{benign setting}. Cronus incurs only a slight degragation in accuracy compared to existing  algorithms, while improves significantly over stand-alone training.
}

\label{fig:benign_convergence}
\end{figure*}

\subsection{Additional Details of the Experiment}
\label{appdx:exp_details}

In this section, we provide the experimental details omitted in Section \ref{exp}.

\paragraphb{Complete robustness assessment.}
In Table \ref{tab:fedlearn} of Section \ref{exp:robustness}, for each dataset and each aggregation algorithm, we showed the accuracy of the attack that is strongest among all the attacks discussed in Section \ref{attacks}.
We compute empirical robustness of aggregation algorithms using this strongest attack as described in Section \ref{setup:metrics}.
Here, in Table \ref{tab:fedlearn_all}, we give the complete evaluation of all the attacks on all of the aggregation algorithms and datasets we consider in this work.
The `Accuracy (Benign)' row of each dataset shows the results in the absence of adversary.
The worst accuracy for a combination of aggregation algorithm and dataset is highlighted in the corresponding column; the corresponding strongest attack can be found from the label of the row of the highlighted cell.
Observe that, label flip attack seems to have lower effect on mean aggregation than the other aggregations; this is because, unlike other aggregations, in case of mean, there is only single malicious client.
Note that, $\mathsf{MWUAvg}$ and $\mathsf{MWUOpt}$ aggregations are robust against all the existing attacks in the literature, but are completely ineffective against the attack we introduced in Section \ref{attack:our_attack}.
Also, note that, Bulyan aggregation is empirically the most robust aggregation after Cronus, but it allows only 25 - 33\% malicious clients compared to other aggregation algorithms such as Krum, in other words, Bulyan has a very low breaking point.
The numbers of malicious parties used in each of our experiments are given in Table \ref{tab:num_parties}.

\paragraphb{Convergence plots for benign setting.}
In Figure \ref{fig:robustness_comparison} of Section \ref{exp}, we show the convergence plots of various aggregation algorithms in federated learning in the adversarial setting.
Figure \ref{fig:benign_convergence} shows the similar convergence plots for the benign setting.
Cronus incurs only a slight reduction in accuracy at a significantly higher gain in robustness and privacy as shown in Sections \ref{exp:robustness} and \ref{exp:membership_risk}.

\end{document}